\documentclass[lettersize,journal]{IEEEtran}
\usepackage{amsmath,amsfonts}
\usepackage{algorithmic}
\usepackage{algorithm}
\usepackage{array}
\usepackage[caption=false,font=normalsize,labelfont=sf,textfont=sf]{subfig}
\usepackage{textcomp}
\usepackage{stfloats}
\usepackage{url}
\usepackage{verbatim}
\usepackage{graphicx}
\usepackage{cite}
\usepackage{hyperref}
\usepackage{xcolor}
\usepackage{pifont}
\begin{document}

\title{ScaleMPA: Rethinking Scalable RRT* Acceleration With a Grid-Native Representation}

\author{Zilong Wang, Yuzhou Chen, Xinyue He, Chen Zhang, Guanghui He*
}



\maketitle

\begin{abstract}
Real-time motion planning remains challenging in large and high-dimensional environments. Prior acceleration of RRT* follows tree-centric state organization, which reduces per-query cost but preserves superlinear end-to-end complexity and limits parallelism through structural dependencies. This paper presents \textbf{ScaleMPA}, a motion-planning accelerator that rethinks RRT* with a \textbf{grid-native representation}. By replacing hierarchical traversal with direct grid-based access, ScaleMPA reduces the planner critical path and exposes fine-grained parallelism. To make this reformulation practical under sparse high-dimensional planning, ScaleMPA further proposes a multi-resolution grid search engine and a hash-grid memory system. Implemented in 28\,nm CMOS, ScaleMPA achieves millisecond-level planning latency and delivers 4.7$\times$--44.4$\times$ speedup over state-of-the-art motion-planning accelerators. 
\end{abstract}

\begin{IEEEkeywords}
Motion Planning, RRT*, Acceleration, Domain-specific Architecture, Algorithm/hardware co-design.
\end{IEEEkeywords}

\section{Introduction}
\label{sec:intro}

\IEEEPARstart{M}{otion} planning is a fundamental component of robotic systems, responsible for generating collision-free trajectories that connect a start state to a goal state under geometric and kinematic constraints~\cite{orthey2023sampling}. It remains indispensable in low-level autonomy stacks for drones, autonomous vehicles, and robotic manipulators, where planning decisions must be made under tight latency budgets and strict safety requirements~\cite{xie2022distributed,safaoui2024safe,orthey2023sampling,de2024humanoid}. As robotic platforms are deployed in increasingly large, cluttered, and high-dimensional environments, the computational cost of motion planning grows rapidly, turning planning latency into a critical system bottleneck rather than a secondary software concern~\cite{orthey2023sampling,de2024humanoid}.

\begin{figure}[!t]
\centerline{\includegraphics[width=0.45\textwidth]{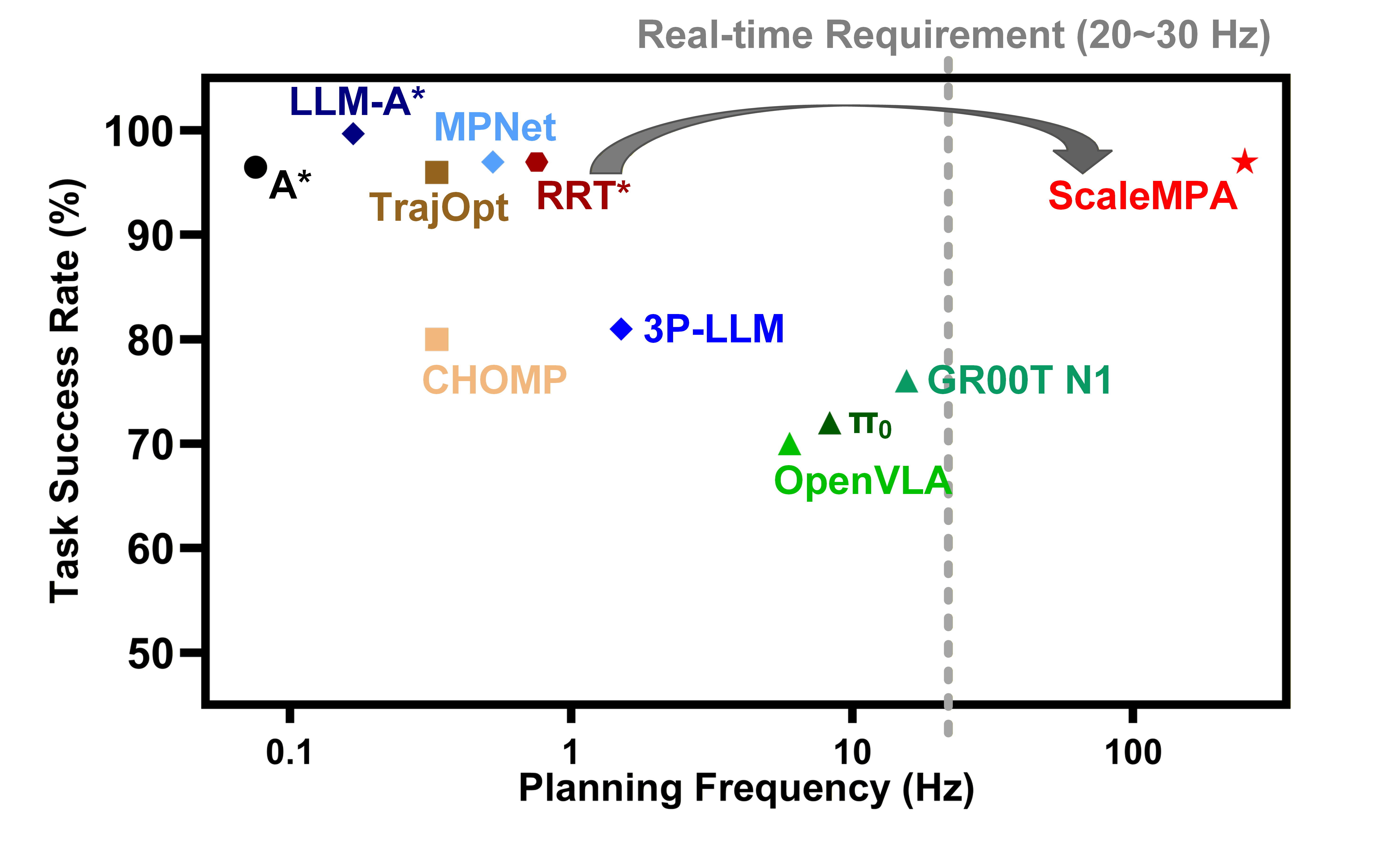}}
\caption{Positioning of ScaleMPA in the motion-planning landscape.} 
\label{img:intro}
\end{figure}

Among existing planning paradigms, sampling-based methods, especially Rapidly-exploring Random Tree Star (RRT*), are attractive to robotic motion planning because they operate directly in continuous configuration spaces and provide strong reliability through explicit geometric reasoning~\cite{hart1968formal,karaman2011sampling,ratliff2009chomp,schulman2014motion}. Learning-based methods, including neural motion planners and LLM-guided planning, can provide useful heuristics or candidate trajectories that improve search efficiency~\cite{qureshi2019motion,latif20243p,meng2024llm}. More recently, vision-language-action (VLA) models have pushed robotic autonomy toward higher-level semantic reasoning and task execution~\cite{kim2024openvla,black2024pi_0,bjorck2025gr00t}. However, these newer approaches do not eliminate the need for classical motion planning. They either operate at a different level of abstraction or do not guarantee exact geometric feasibility and robustness by themselves, which remain essential in low-level, safety-critical settings. As illustrated in Fig.~\ref{img:intro}, this paper focuses on precisely this regime, where the central challenge is to retain the reliability of classical planning while substantially improving its execution efficiency.


\begin{figure*}[!t]
\centerline{\includegraphics[width=\textwidth]{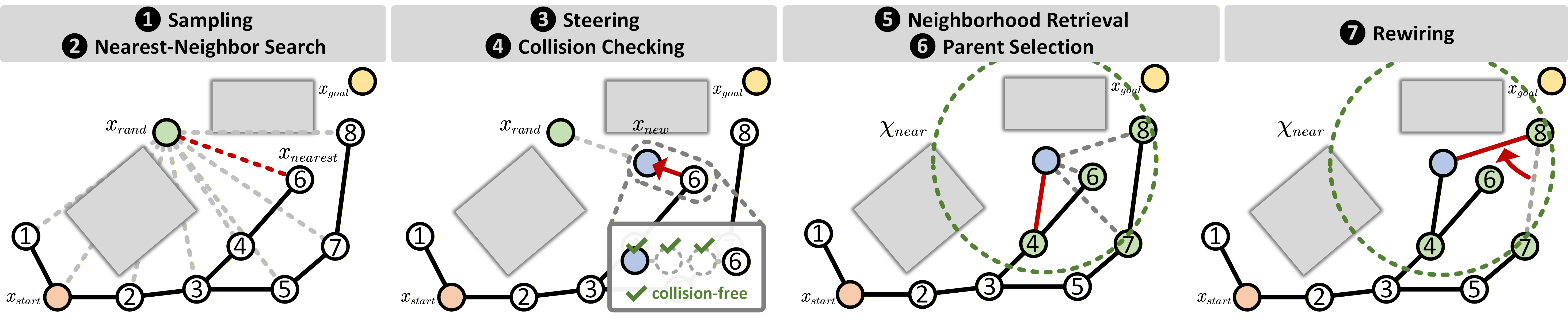}}
\caption{An illustration of one iteration of the RRT* algorithm. }
\label{img:background}
\end{figure*}

Despite its algorithmic appeal, RRT* scales poorly in large environments and high-dimensional configuration spaces. Standard RRT* performs $\Theta(n)$ exploration iterations, where $n$ denotes the number of explored nodes. In naive implementations, each iteration scans the explored nodes for neighbor search and the obstacle set for collision checking, leading to $\Theta(n^2+Mn)$ total work, where $M$ is the number of obstacles. Even with aggressive software optimization, planning latency can grow to tens or even hundreds of seconds on modern CPUs, far beyond the 20--30\,Hz planning rate required by dynamic robotic systems~\cite{strub2020advanced,ortiz2024idb,fly2019icra,replan2027iros,ionescu2021adaptive}. This gap has motivated a series of hardware acceleration efforts. 
\IEEEpubidadjcol

Prior accelerators reduce query cost with octrees, R-trees, and BVHs~\cite{dac2018dadup,isca2023mp,hpca2023parallelnn,ijrr2012gpucc,hpca2024moped,hpca2020quicknn,sig2022rtnn}, but retain \textbf{tree-centric} state organization.
As a result, the design space is constrained by two coupled bottlenecks. First, although a single query can be reduced from $\Theta(n)$ to $\Theta(\log n)$, the end-to-end complexity of RRT* remains $\Theta(n \log n)$ because every iteration still performs multiple queries and updates on the hierarchy. Second, tree updates are structurally dependent: insertions and revisions contend on shared ancestors, require backtracking, and introduce synchronization hotspots that are unfriendly to parallel hardware. 

This paper argues that breaking this wall requires rethinking the representation of planner state, rather than only accelerating tree traversal. To this end, we propose \textbf{ScaleMPA}, a scalable motion-planning accelerator that reformulates RRT* around a \textbf{grid-native representation}. Instead of organizing obstacles and explored nodes in hierarchical trees, ScaleMPA directly maps the configuration space into regularly indexed cells. This change has two architectural consequences. First, it removes root-to-leaf traversal from the critical path, reducing the average cost of accessing local planning context and
exhibiting near-linear end-to-end scaling across the evaluated workloads. Second, it exposes naturally independent units of work across cells, enabling fine-grained parallel query and update operations without the parent-node conflicts that limit tree-based designs.



However, simply replacing a tree with a uniform grid is not sufficient. A naive grid introduces two major obstacles that negate its theoretical benefit. (1) \textbf{worst-case query degradation}: in sparse, high-dimensional spaces, many neighbor-search queries land in empty cells, forcing fallback scans that can drive the effective complexity back toward $\Theta(n^2)$. In our profiling, more than 96\% of queries in a 7-DOF planning task hit empty cells, leading to a worst-case latency of 465\,ms, which is 5$\times$ slower than a SOTA tree-based accelerator. (2) \textbf{prohibitive memory footprint}: unlike a tree that stores only occupied regions, a dense grid must reserve space for the full configuration space, causing severe storage blowup. In the same high-dimensional setting, a naive grid requires 84,332\,KB, which is over three orders of magnitude larger than a tree-based alternative. These two issues show that the real challenge is not merely identifying the opportunity of a grid, but making a grid-based planner representation practical for sparse high-dimensional motion planning.

At the same time, removing the tree-centric bottleneck does not make GPUs a suitable execution platform. Even with a grid-native reformulation, real-time motion planning remains fundamentally mismatched to commodity GPU architectures because it continues to exhibit irregular control flow, data-dependent search expansion, sparse and non-coalesced memory access, and severe thread-level workload imbalance. In particular, neighbor search remains highly irregular under empty-cell misses, while rewiring and commit are latency-sensitive and difficult to sustain efficiently on throughput-oriented GPU architectures. Consequently, although GPUs can accelerate selected kernels, they remain ill-suited for efficient and predictable end-to-end real-time planning. This mismatch motivates a dedicated accelerator.

ScaleMPA addresses these challenges through an algorithm--architecture co-design. To mitigate query-miss penalties, we develop a \textbf{multi-resolution grid search engine} that dynamically selects grid granularity according to planning density, using coarse grids when exploration is sparse and finer grids when local neighborhoods become denser. This design reduces the worst-case latency in the 7-DOF setting from 465\,ms to 5\,ms. To tame storage blowup, we design a \textbf{hash-grid memory system} that compresses sparse logical grids into compact physical storage by allocating memory only for occupied cells and exploiting the sparsity of high-dimensional planning spaces. This reduces storage overhead to 105.9\,KB, small enough to fit on-chip and avoid costly random off-chip access. Together, these mechanisms turn a theoretically attractive but impractical grid formulation into a hardware-efficient planning substrate. Built on top of this representation, ScaleMPA further introduces a pipelined and parallel planning architecture that supports neighbor search, collision checking, and tree refinement in an end-to-end hardware dataflow. 

\begin{figure*}[!t]
\centerline{\includegraphics[width=1.05\textwidth]{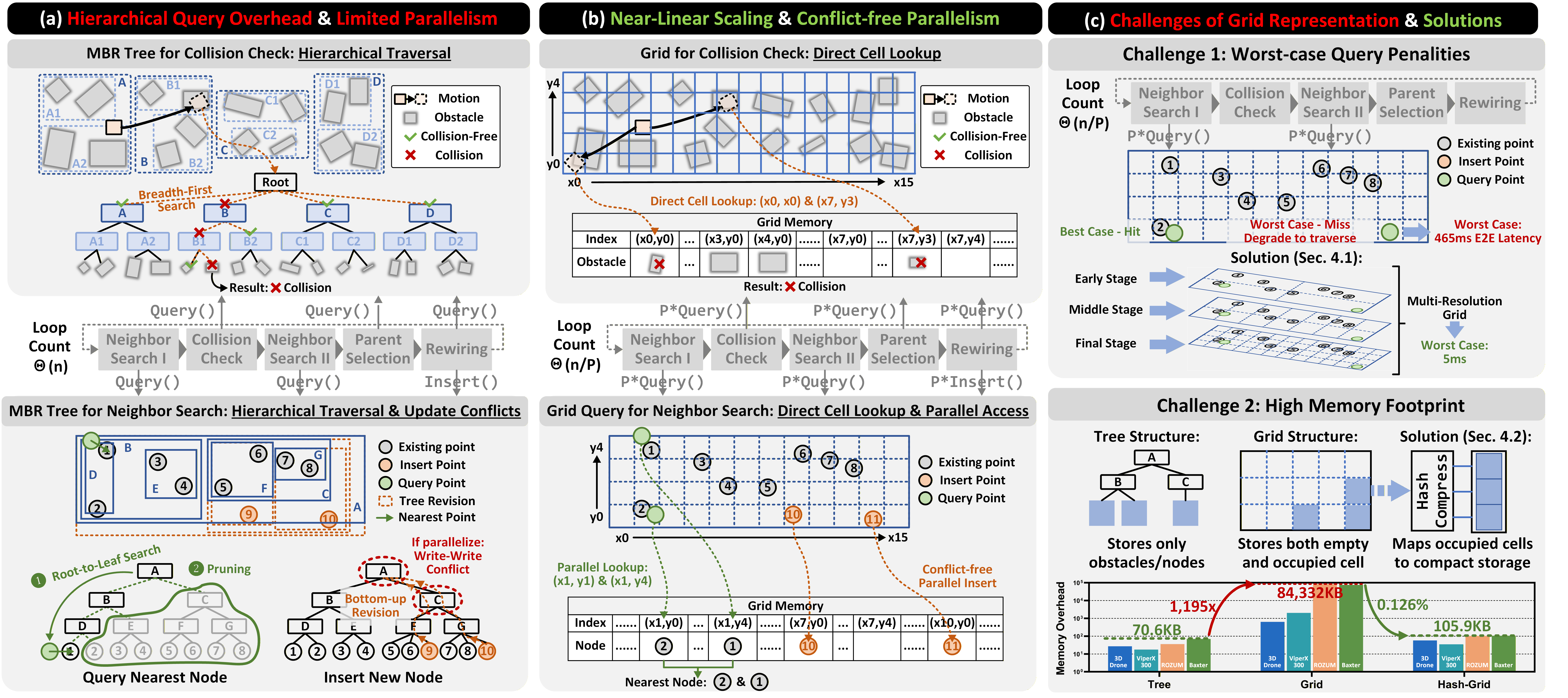}}
\caption{Motivation for a grid-based representation. (a) Tree-based methods incur hierarchical traversal overhead and limited parallelism. (b) The proposed grid-based approach reduces traversal overhead and enables higher parallelism. (c) Challenges to applying a grid-based representation.}
\label{img:motivation1}
\end{figure*}

In summary, this paper makes the following contributions:
\begin{itemize}
    \item We identify that the key scalability bottleneck of prior RRT* accelerators lies in their \textbf{tree-centric representation}, which preserves $\Theta(n \log n)$ end-to-end complexity and introduces structural dependencies that limit hardware parallelism.
    \item We propose a \textbf{grid-native reformulation} of RRT* that replaces hierarchical traversal with direct cell-based access, exposing conflict-free fine-grained parallelism and exhibiting near-linear end-to-end scaling across the evaluated workloads.
    \item We propose a \textbf{multi-resolution query mechanism} and a \textbf{compressed hash-grid memory system}, jointly overcoming empty-cell miss penalties and storage explosion issues for sparse high-dimensional space.
    \item We realize these ideas in \textbf{ScaleMPA}, an end-to-end motion-planning accelerator that delivers substantial speedup, area efficiency, and energy efficiency improvements over prior software and hardware baselines.
\end{itemize}
\section{Background and Motivation}
\label{sec:background}

\subsection{RRT* and Its Dominant Operators}
\label{subsec:rrt_background}

RRT* is a representative sampling-based planner that incrementally constructs a search tree in continuous configuration space while preserving asymptotic optimality~\cite{karaman2011sampling}. Each iteration performs sampling, nearest-neighbor search, steering, collision checking, neighborhood retrieval, parent selection, and rewiring, as illustrated in Fig.~\ref{img:background}.

\ding{182} \textit{Sampling.}  
A random configuration $x_\text{rand}$ is sampled from the configuration space $\mathcal{C}$, representing a valid robot state defined by its position, orientation, or joint angles. Sampling is typically uniform but may be biased toward the goal to accelerate convergence.

\ding{183} \textit{Nearest-Neighbor Search.} After sampling $x_\text{rand}$, RRT* identifies the nearest node $x_\text{nearest}$ in the current exploration tree $\mathcal{T}$ using a distance metric.

\ding{184} \textit{Steering.}  
The algorithm generates a new node $x_\text{new}$ by extending from $x_\text{nearest}$ toward $x_\text{rand}$ with a fixed or bounded step size, subject to the robot's motion constraints.

\ding{185} \textit{Collision Checking.}  
Before adding $x_\text{new}$ to the tree, the algorithm verifies that the path from $x_\text{nearest}$ to $x_\text{new}$ is collision-free. The robot's geometry is swept along the path, and $Q$ discrete points $\{x_{\text{col}:0}, x_{\text{col}:1}, \dots, x_{\text{col}:Q-1}\}$ are checked using the \textit{Separating Axis Theorem}. If a collision is detected, the iteration terminates and $x_\text{new}$ is discarded.

\ding{186} \textit{Neighborhood Retrieval.}  
The algorithm retrieves all existing nodes $\mathcal{X}_\text{near} = \{x_\text{near:0}, x_\text{near:1}, \dots, x_\text{near:K-1}\}$ within a radius $r_\text{near}$ of $x_\text{new}$ as candidates for rewiring.

\ding{187} \textit{Parent Selection.}
The node $x_{\text{min}}$ that minimizes the cost to $x_\text{new}$ via a collision-free path is selected from $\mathcal{X}_\text{near}$. After validating the connection, $x_{\text{min}}$ becomes the parent of $x_\text{new}$.

\ding{188} \textit{Rewiring.}  
After inserting $x_\text{new}$ as a child of $x_\text{min}$, the algorithm performs local rewiring. For each $x_\text{near} \in \mathcal{X}_\text{near}$, it checks whether connecting through $x_\text{new}$ reduces the path cost while preserving collision-free connectivity. If so, the parent of $x_\text{near}$ is updated to $x_\text{new}$ after validating the new connection.

Although the computation within a single expansion step is modest, overall runtime is dominated by repeated accesses to a dynamically growing planner state. These accesses are concentrated in two operator classes. The first is \emph{geometric filtering}, primarily collision checking, which validates candidate motions against obstacles. The second is \emph{neighborhood discovery and connectivity update}, which identifies nearby nodes and updates local tree structure through parent selection and rewiring. As exploration proceeds, both classes are invoked repeatedly, making RRT* increasingly sensitive to how planner state is organized and accessed. The key implication is that the scalability of RRT* is determined not only by the cost of geometric primitives, but also by the representation used to organize planner state.

\subsection{Tree-based Method Hits a Scalability Wall}
\label{subsec:tree_wall}

To reduce exhaustive scans, prior accelerators organize planner state hierarchically using octrees, R-trees, KD-trees, and BVHs~\cite{dac2018dadup,isca2023mp,dac2020dadu,hpca2023parallelnn,ijrr2012gpucc,hpca2024moped,hpca2020quicknn,sig2022rtnn}. These structures cluster sampled nodes or obstacles into nested regions and prune irrelevant branches during neighbor search and collision checking, substantially reducing the cost of individual queries.

However, this line of work retains a tree-centric execution model and therefore inherits two fundamental limitations. First, as illustrated in Fig.~\ref{img:motivation1}(a), hierarchical pruning improves per-query complexity but does not eliminate end-to-end superlinear growth. Even if a query is reduced from $\Theta(n)$ to roughly $\Theta(\log n)$, RRT* still performs $\Theta(n)$ iterations, each containing multiple searches and updates, leaving the overall complexity at $\Theta(n \log n)$. The limitation is not the effectiveness of pruning, but the fact that its benefit remains local to each query.

\begin{figure}[!t]
\centerline{\includegraphics[width=0.4\textwidth]{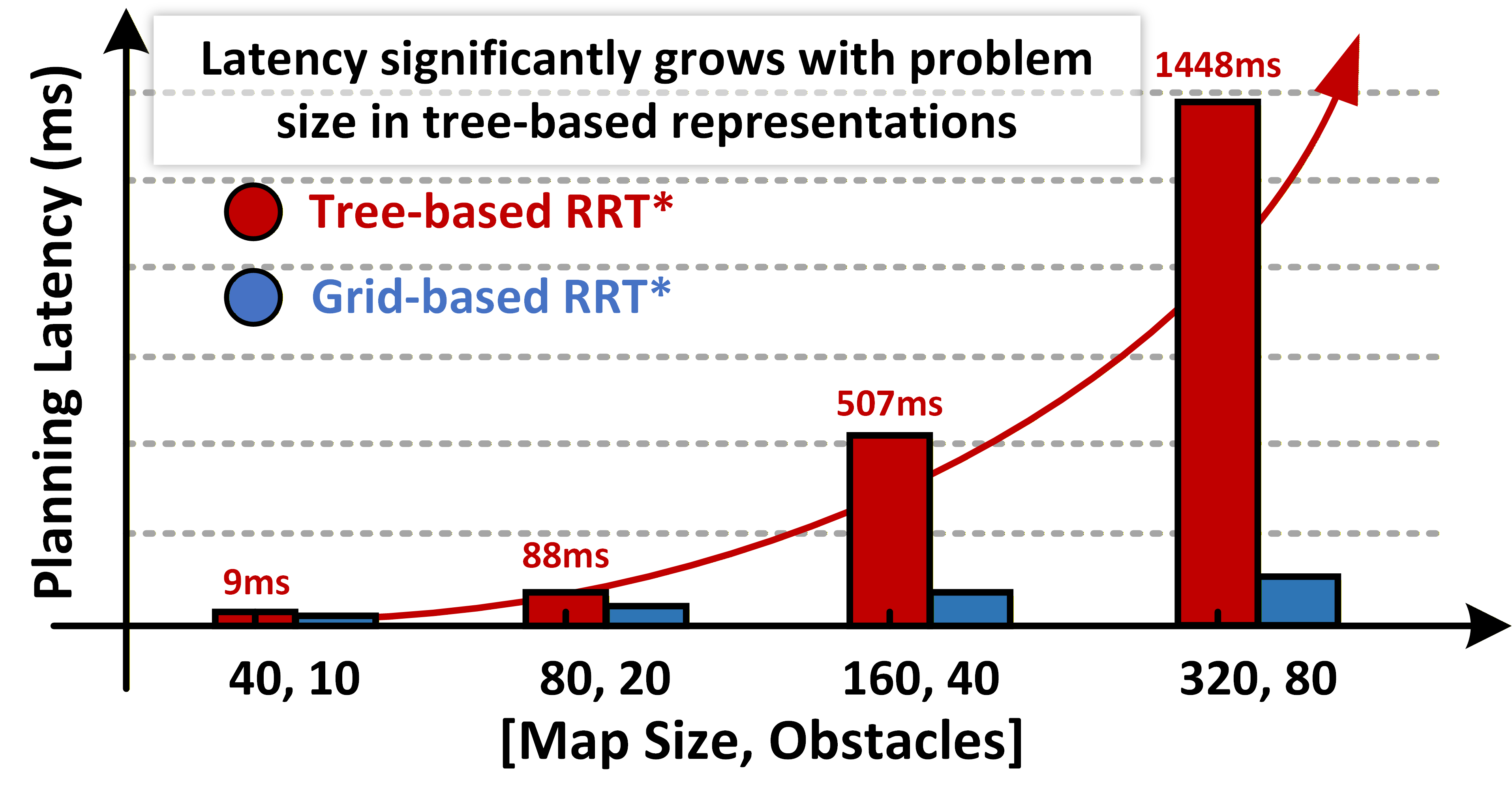}}
\caption{Scalability comparison between tree-based and grid-based RRT*. }
\label{img:scalability_wall}
\end{figure}

Second, tree-based representations are structurally unfriendly to parallel execution. As shown in Fig.~\ref{img:motivation1}(a), root-to-leaf traversal induces sequential dependence, while insertions, rewiring, and region updates contend on shared ancestors. Tree imbalance further amplifies divergence across concurrent queries. In hardware, these effects manifest as synchronization hotspots, update conflicts, and serialization around shared nodes. Consequently, prior work accelerates RRT* \emph{within} a tree abstraction, but does not remove the abstraction itself as the source of the scalability wall. Fig.~\ref{img:scalability_wall} further confirms this limitation by showing rapidly increasing latency as the planning problem scales.

\subsection{A Grid-native Representation Is Promising}
\label{subsec:grid_opportunity}

A grid-native representation provides a natural alternative. Rather than recursively organizing planner state, a grid partitions configuration space into regularly indexed cells, each corresponding to a local spatial region. It has two advantages:

First, a grid replaces traversal-dependent lookup with direct indexing. In a tree, query latency depends on a data-dependent root-to-leaf walk. In a grid, coordinates map directly to cells, allowing relevant local context to be accessed without hierarchical traversal, as illustrated in Fig.~\ref{img:motivation1}(b). This replaces pointer-chasing with a more regular access pattern and provides the basis for near-linear average planner scaling when local grid queries remain bounded.

Second, a grid exposes a more hardware-friendly unit of parallelism. In a tree, multiple operations conflict when they touch overlapping paths in the hierarchy. In a grid, accesses to different cells are naturally decoupled, substantially weakening structural dependence across concurrent operations, as illustrated in Fig.~\ref{img:motivation1}(b). This creates a better foundation for fine-grained parallelism in neighbor search, collision checking, and node insertion. In this sense, a grid is not merely a different index; it is a representation-level shift that changes both planner-state access and the granularity of parallel work.


\subsection{Limitations of the Naive Grid Formulation}
\label{subsec:naive_grid_limits}

The representation alone, however, is insufficient. It suffers from two fundamental limitations, as illustrated in Fig.~\ref{img:motivation1}(c): The first is \emph{sparsity-induced miss penalty}. In large or high-dimensional spaces, occupancy is highly sparse, so many neighbor-search queries fall into empty cells and must expand outward to locate valid nearby nodes. In the worst case, this fallback behavior reintroduces near-exhaustive traversal and can drive effective complexity back toward $\Theta(n^2)$. This reflects a sparsity--resolution tradeoff: finer grids improve selectivity but also increase the probability of empty-cell misses.

The second is \emph{mismatch between logical regularity and physical sparsity}. A dense grid is logically regular, but the planner state is physically sparse. Allocating storage for the full logical space therefore wastes substantial memory on empty cells, particularly in high-dimensional settings. This can make on-chip storage infeasible and force costly off-chip accesses, undermining the low-latency advantage that motivated the grid representation in the first place.

These limitations show that exposing grid-level parallelism is not enough; the representation must also be made robust to sparsity in both query execution and storage organization.

\subsection{Commodity GPUs Remain Insufficient}
\label{subsec:why_not_gpu}

Although prior GPU-parallel and vectorized software planners have substantially accelerated specific sampling-based formulations~\cite{bialkowski2011massively,thomason2024motions,
wilson2025nearest,huang2025prrtc}, a grid-native formulation, despite
exposing substantially greater regularity and parallelism than
tree-based RRT*, does not by itself make commodity GPUs a sufficient
platform for real-time RRT* execution. Representation-level parallelism
does not automatically translate into efficient end-to-end execution
on a throughput-oriented architecture.


\begin{figure}[!t]
\centerline{\includegraphics[width=0.5\textwidth]{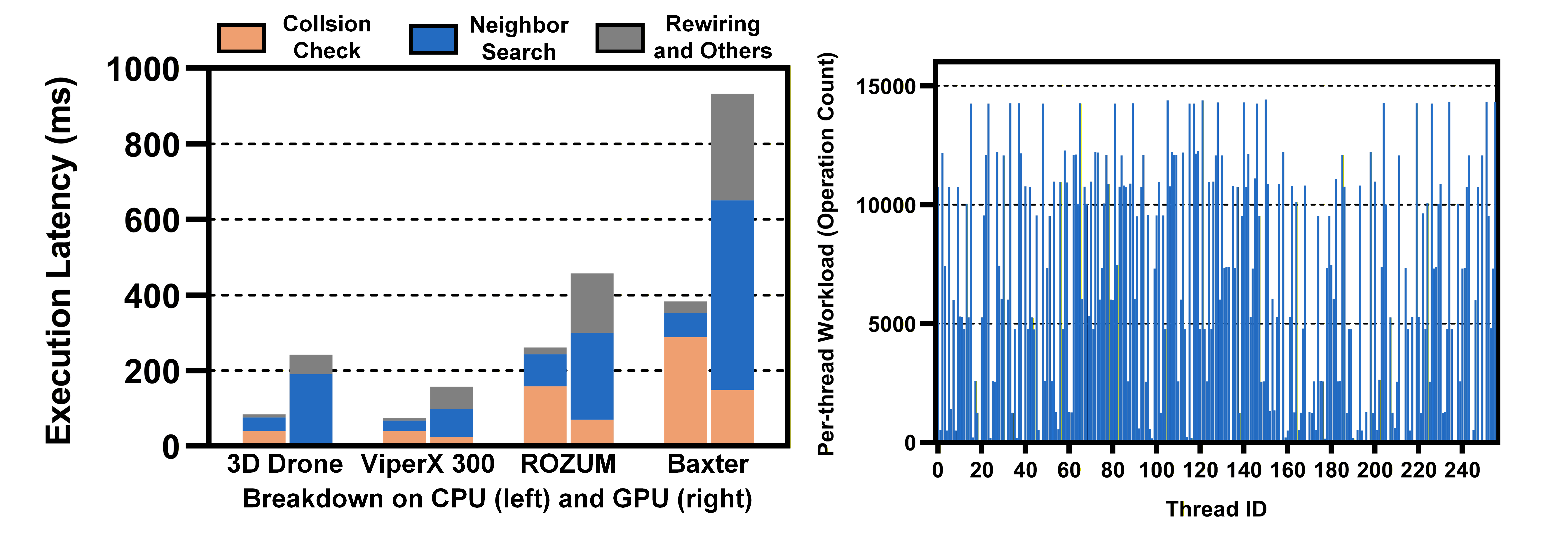}}
\caption{Execution Latency Breakdown and Thread-Level Workload Distribution. (a) GPU speedup varies across operations. (b) Thread-level workload imbalance in neighbor search limits GPU efficiency. }
\label{img:eval_gpu}
\end{figure}

\begin{table}[!t]
\caption{Summary of warp memory statistics\label{tab:eval_gpu_mem}}
\centering
\scriptsize
\begin{tabular}{|c||c|c|c|}
\hline
\shortstack{\textbf{Operation} \\ \textbf{Stage}}
& \shortstack{\textbf{Mean Unique} \\ \textbf{Cache Lines}}
& \shortstack{\textbf{Mean Address} \\ \textbf{Span}}
& \shortstack{\textbf{Non-coalesced} \\ \textbf{Ratio}} \\
\hline
Collision Check      & 1.947   & 216.154          & 0.000 \\
\hline
Neighbor Search      & 20.578  & $2.55\times10^6$ & 0.649 \\
\hline
Rewiring and Others  & 0.819   & 19.223           & 0.000 \\
\hline
\end{tabular}
\end{table}

As shown in Fig.~\ref{img:eval_gpu}(a), GPU acceleration varies significantly across planner stages. Collision checking benefits from relatively regular computation and thus achieves more stable speedup. However, both neighbor search and rewiring-related stages remain major latency contributors, as they exhibit inherently irregular and data-dependent execution patterns that are poorly matched to throughput-oriented architectures. Moreover, GPU workloads are highly imbalanced across threads, particularly during the neighbor search stage, as shown in Fig.~\ref{img:eval_gpu}(b). The number of operations per thread varies by orders of magnitude, due to data-dependent expansion during empty-cell misses and spatially non-uniform node distributions. This imbalance leads to severe warp divergence and underutilization of SIMD lanes, limiting effective parallelism.



Table~\ref{tab:eval_gpu_mem} further reveals poor memory efficiency in GPU execution. Neighbor search exhibits a large address span and a high non-coalesced access ratio, indicating that memory accesses are both sparse and irregular. Although other stages exhibit relatively efficient memory behavior, the planner operates as a tightly coupled pipeline, and the inefficiency of neighbor search ultimately limits the overall execution efficiency.

\subsection{Design Implications}
\label{subsec:design_implications}

The above analysis leads to four design implications. First, the critical path of the planner should avoid hierarchical traversal and instead favor direct state access. Second, the representation should expose conflict-free or low-conflict units of parallel work so that concurrency is not bounded by shared structural dependencies. Third, the execution substrate must tolerate sparsity without suffering miss-driven latency collapse. Fourth, the storage system should match sparse occupancy rather than dense logical space, allowing planner state to remain compact and on chip whenever possible. These implications suggest that the overall design should optimize for end-to-end low-latency execution rather than only high operator throughput.

These observations motivate \textbf{ScaleMPA}, which co-designs planner representation, query execution, and storage organization to achieve scalable real-time motion planning. In the next section, we present the algorithmic foundation of ScaleMPA and show how a grid-native formulation can be made hardware-efficient.

\begin{figure*}[!t]
\centerline{\includegraphics[width=0.95\textwidth]{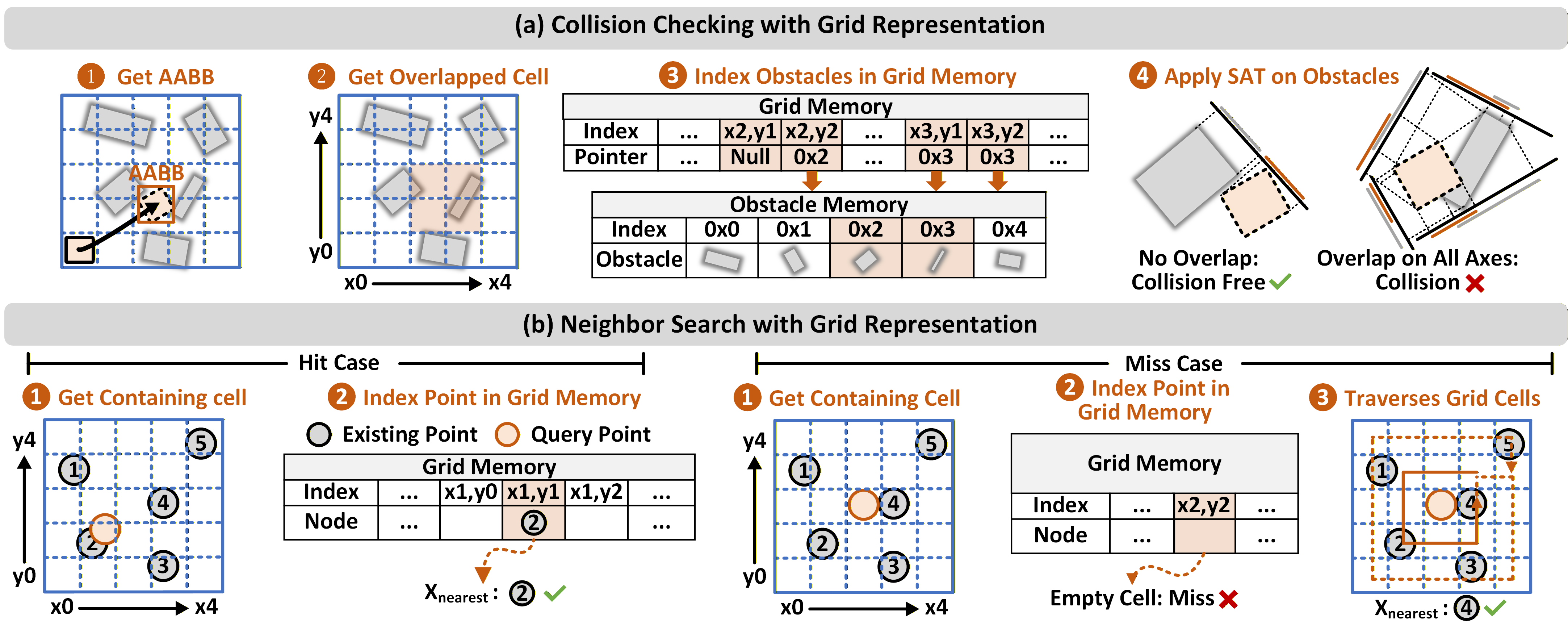}}
\caption{ Key operators with grid representation. (a) Collision checking. (b) Neighbor search.}
\label{img:algorithm_overall}
\end{figure*}

\section{ScaleMPA: Algorithm Design}


This section presents the algorithmic foundation of \textbf{ScaleMPA}, which rethinks RRT* through the lens of a grid-native representation. Our design is driven by two primary objectives: (1) reducing the average cost of the dominant operators, including collision checking and neighbor search, and (2) unlocking fine-grained parallelism across iterations. To address these two objectives separately, Section~\ref{sec:key_operator} introduces the grid-native reformulation of key operators, while Section~\ref{sec:parallel_rrt} presents a conditional parallel algorithm that overcomes cross-iteration dependencies.


\subsection{Optimizing Key Operators with Grid Access} 
\label{sec:key_operator}

\subsubsection{Grid-based Collision Checking}

Collision checking is a dominant bottleneck in RRT*, as each candidate path segment must be verified against obstacles using the computationally intensive Separating Axis Theorem (SAT). In 3D, a single SAT test requires 459 multiplications and 315 additions, so repeated SAT verification can dominate the total computation in cluttered environments.

To eliminate redundant SAT invocations, we introduce a grid-based pre-filtering mechanism that strategically prunes distant obstacles. As illustrated in Fig.~\ref{img:algorithm_overall}(a), the environment is partitioned into uniform grid cells, each indexed by its spatial centroid and storing contained obstacles in \emph{Oriented Bounding Box (OBB)} format. For a given robot configuration, we first compute its \emph{Axis-Aligned Bounding Box (AABB)} by projecting the robot OBB corners along coordinate axes. Only grid cells overlapping with this AABB are retrieved, and solely the obstacles within these cells undergo SAT testing. This spatial pruning eliminates distant, non-interfering obstacles, dramatically reducing computational overhead.

\textbf{Complexity and Correctness Analysis.}
Exhaustive collision checking requires $\Theta(M)$ SAT tests per local collision query, where $M$ is the number of obstacles. Tree-based broad-phase methods reduce this to $\Theta(\log M)$ traversal under a balanced obstacle tree and bounded candidate verification. Grid-based pre-filtering replaces hierarchical traversal with direct cell access. Its query cost is determined by the number of overlapped cells and local obstacle candidates rather than
the global obstacle count $M$. In our evaluated environments, the grid filter returns only 0.91 SAT candidates on average per robot OBB at each sampled collision-checking point, with 90th/99th percentiles of 4/10. Thus, the common-case collision-checking cost is nearly constant with respect to $M$.


Importantly, this reduction does not compromise collision-checking correctness. The grid only filters distant obstacles, and all remaining candidates are still verified by SAT, so collision validity is unchanged.

\subsubsection{Grid-Based Neighbor Search} 

Neighbor search constitutes the second major bottleneck, involving: (1) finding the nearest node $x_{\text{nearest}}$ to a random sample $x_{\text{rand}}$, and (2) retrieving the neighborhood $\mathcal{X}_{\text{near}}$ within radius $r$.

\begin{table}[!t]
\caption{Approximation accuracy of grid-based neighbor search\label{tab:nn_approximation}}
\centering
\begin{tabular}{|c||c|c|}
\hline
\shortstack{\textbf{Planning Scenario}}
& \shortstack{\textbf{Exact Match Rate}}
& \shortstack{\textbf{Avg. Distance Increase}} \\
\hline
3D Drone      & 93.1\%   & 1.7\%\\
\hline
ViperX 300    & 92.6\%   & 1.0\%\\
\hline
ROZUM         & 91.9\%   & 0.9\%\\
\hline
Baxter        & 92.4\%   & 1.1\%\\
\hline
\end{tabular}
\end{table}

Tree-based approaches locate candidate nodes by traversing a hierarchical index. In contrast, the grid-based formulation directly maps spatial coordinates to grid cells containing explored nodes. As shown in Fig.~\ref{img:algorithm_overall}(b)-step\,\ding{183}, the query point first identifies its containing cell. In a \emph{hit} scenario, the search completes in one access, returning the nearest candidate within that cell as $x_{\text{nearest}}$. For \emph{miss} cases (empty target cell), adjacent cells are scanned in expanding Manhattan distance until non-empty cells are found. All retrieved nodes are then compared by distance, with the nearest node (e.g., Node-4 in Fig.~\ref{img:algorithm_overall}(b)) selected as the final result.

\textbf{Complexity and Approximation Analysis.}
For nearest-neighbor search, a naive implementation requires $\Theta(n)$ distance evaluations over all explored nodes, while balanced tree-based indexes reduce this cost to $\Theta(\log n)$. Grid-based search replaces hierarchical traversal with direct cell indexing and requires only constant local work when the query hits an occupied cell with bounded local candidates. However, its actual cost depends on empty-cell misses and the resulting search expansion. As discussed in Section~\ref{subsec:naive_grid_limits}, naive single-resolution grids may suffer from sparsity-induced outward expansion and degrade toward exhaustive traversal. This motivates the multi-resolution grid search engine in Section~\ref{sec:multi_resolution}, which reduces empty-cell misses and limits search expansion.

Regarding approximation, the returned $x_{\text{nearest}}$ is the nearest node within the retrieved grid candidate set, not necessarily the exact global nearest node. Table~\ref{tab:nn_approximation} shows that it matches the exact nearest neighbor in 92.5\% of queries, and the remaining mismatches increase the nearest-neighbor distance by only 1.2\% on average. This indicates a high-match-rate local approximation with small distance error.

\subsubsection{Grid Size Selection}

Grid size critically balances memory efficiency and operational performance. An overly fine grid increases traversal overhead, while an excessively coarse grid reduces filtering effectiveness. Empirical analysis establishes optimal grid sizes: approximately matching robot OBB dimensions for collision checking, and 10-20 times the steering stride for neighbor search. This calibration ensures high hit rates while minimizing unnecessary cell visits, achieving an optimal trade-off for scalable performance.

\subsection{Unlocking RRT* Cross-Iteration Parallelism}\label{sec:parallel_rrt}

\begin{figure}[!t]
\centerline{\includegraphics[width=0.5\textwidth]{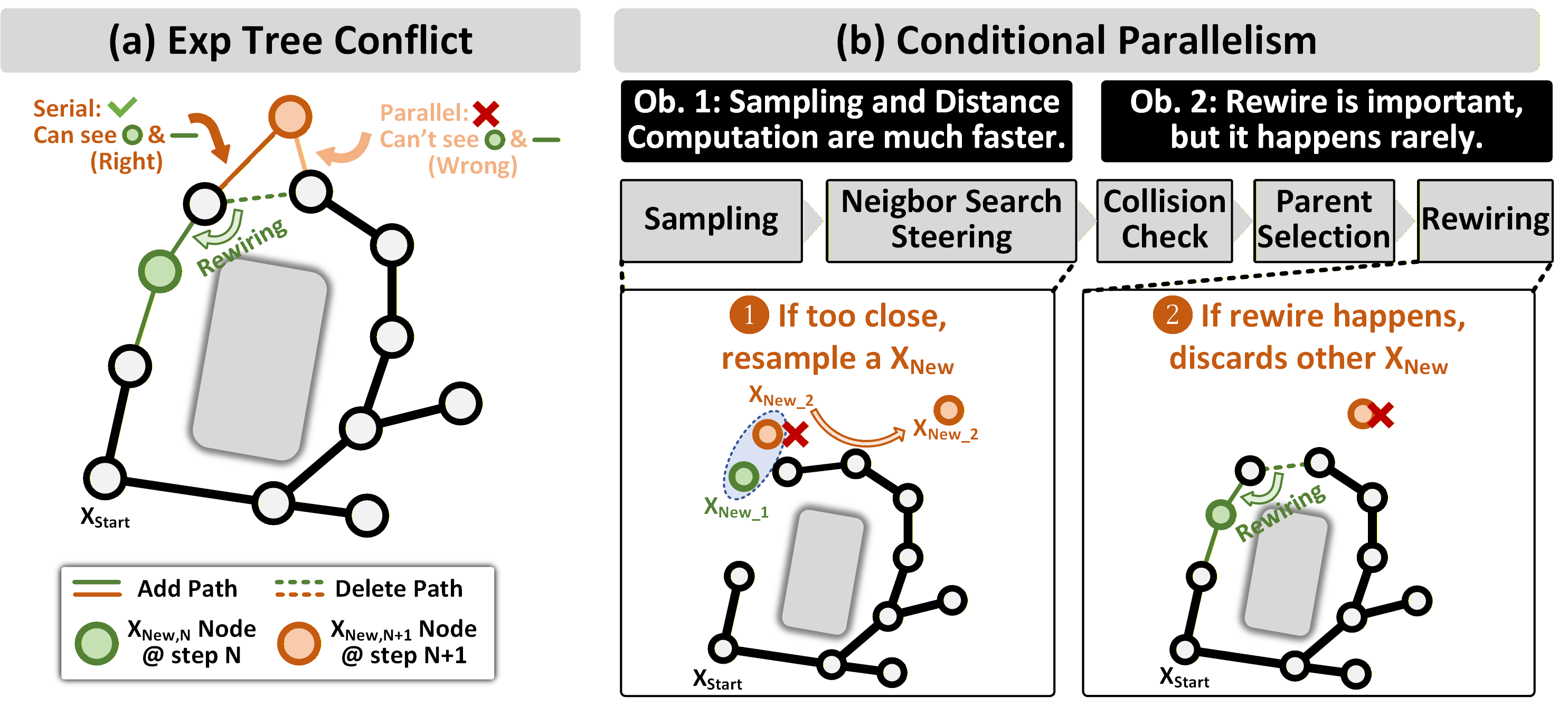}}
\caption{Conditional Parallelism for Accelerating RRT*. (a) Naive parallel processing produces incorrect results during rewiring. (b) Our conditional parallelism employs spatial sampling and speculative execution to maintain correctness while enabling parallel acceleration.}
\label{img:conditional_parallelism}
\end{figure}

Despite operator-level optimizations, RRT* exhibits cross-iteration dependencies that limit parallelism, as illustrated in Fig.~\ref{img:conditional_parallelism}(a): (1) closely located samples may become mutual neighbors, affecting neighbor search results; (2) rewiring operations create read-after-write hazards by modifying parent-child relationships.


\subsubsection{Conflict-free Parallelism via Spatial Sampling} 

We ensure spatial isolation among parallel samples by generating $N$ random configurations and rejecting those falling within overlapping grid cells or violating minimum distance constraints, as illustrated in Fig.~\ref{img:conditional_parallelism}(b). This guarantees independent execution of neighbor search, steering, and collision checking operations. The resampling overhead is negligible, as the samples are produced by simple random number generators implemented with Linear Feedback Shift Registers.

\subsubsection{Speculative Rewiring and Commit Protocol} Rewiring introduces read-after-write hazards when multiple $x_{\text{new}}$ candidates attempt to update the same $x_{\text{near}}$ node. However, benchmark profiling shows that rewiring occurs infrequently ($\sim$ 0.7\% per iteration). We exploit this sparsity through speculative execution, as illustrated in Fig.~\ref{img:conditional_parallelism}(b). All $x_{\text{new}}$ nodes proceed through the planning pipeline. At the rewiring stage, a commit protocol is triggered: If no rewiring is detected, all new nodes are inserted into the tree in parallel. If multiple rewiring attempts conflict, only the earliest valid one is committed. Other speculative updates are discarded to ensure consistency. This \emph{conditional parallelism} maintains algorithmic correctness while enabling speculative tree expansion. Combined with grid-based filtering, it delivers up to $11.3\times$ speedup over tree-based RRT* without sacrificing path quality.

\section{ScaleMPA: Hardware Design}
\label{sec:hw}

Building upon the algorithmic optimizations in Section~\ref{sec:key_operator} and Section~\ref{sec:parallel_rrt}, we design a specialized hardware accelerator, \textit{ScaleMPA}, that fully exploits grid-based RRT* operations. To overcome the \textbf{worst-case query penalty} challenge, Section~\ref{sec:multi_resolution} proposes a \textbf{multi-resolution grid search engine}, which adaptively employs different grid resolutions to avoid empty-cell queries. To reduce the \textbf{high memory footprint}, Section~\ref{sec:hash_memory} describes the proposed \textbf{hash-grid memory} that exploits hash collisions to compress sparse grid storage. Finally, Section~\ref{sec:dataflow_scheduling} details the overall architecture, pipelined dataflow, and parallel scheduling strategy that collectively enable ScaleMPA's high-performance operation.

\subsection{Multi-resolution Grid Search Engine} \label{sec:multi_resolution}

\begin{figure}[!t]
\centerline{\includegraphics[width=0.5\textwidth]{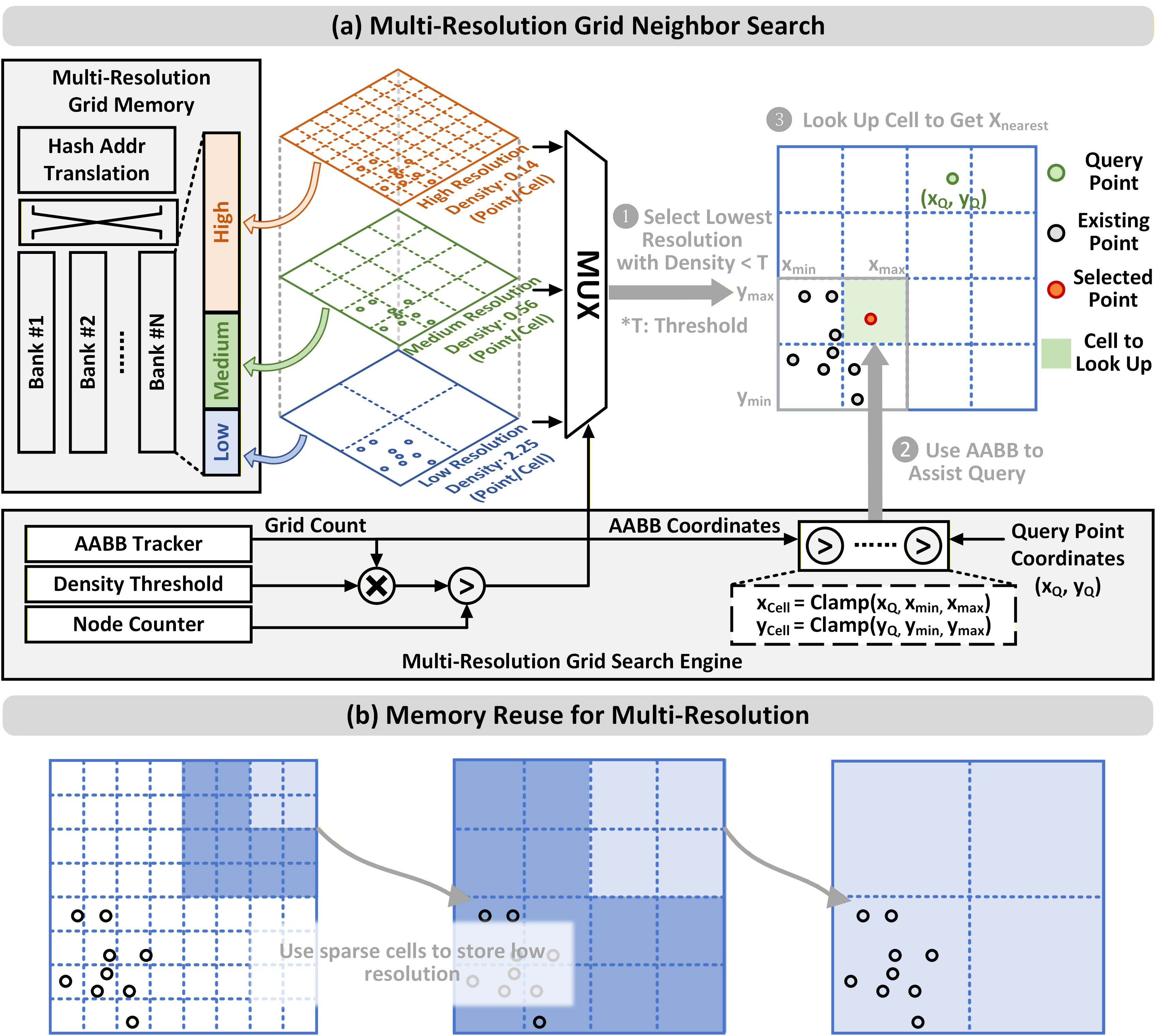}}
\caption{Multi-resolution Grid Search Engine. (a) Multi-resolution search engine selects the solution based on node counter and AABB tracker. The AABB-assisted lookup unit relocates queries to avoid empty-cell accesses. (b) Dynamic memory reallocation releases low-resolution grid memory for reuse in finer grids.}
\label{img:multi_resolution}
\end{figure}

As discussed in Section~\ref{subsec:naive_grid_limits} and Section~\ref{sec:key_operator}, grid-based neighbor search faces significant performance degradation when queries target empty cells, potentially falling back to exhaustive $O(n)$ scans that undermine real-time guarantees. To overcome this fundamental challenge, ScaleMPA incorporates a \textbf{multi-resolution grid search engine} that dynamically adjusts grid resolution based on spatial occupancy patterns, effectively mitigating worst-case penalties without incurring substantial storage overhead.

As illustrated in Fig.~\ref{img:multi_resolution}, each planning instance maintains three grid resolution levels concurrently. A global node counter and an AABB tracker monitor the total number of explored nodes and the spatial extent of the explored region. Based on real-time density metrics, the engine intelligently selects low, medium, or high-resolution grids for neighbor queries:

\begin{itemize}
    \item \textbf{Early Planning Stages}: Sparse node distribution favors low-resolution grids, ensuring most queries target occupied cells and avoiding miss penalties.
    \item \textbf{Late Planning Stages}: Dense node configurations benefit from high-resolution grids, reducing intra-cell search complexity while maintaining high hit rates.
\end{itemize}

To further optimize query efficiency, an AABB-assisted lookup unit relocates queries within the explored region boundaries through clamping operations, significantly increasing hit rates and minimizing fallback traversal. Even when misses occur, coarse-grained grids combined with AABB confinement enable locating valid nodes with minimal empty cell queries rather than exhaustive scans. 

\begin{figure}[!t]
\centerline{\includegraphics[width=0.5\textwidth]{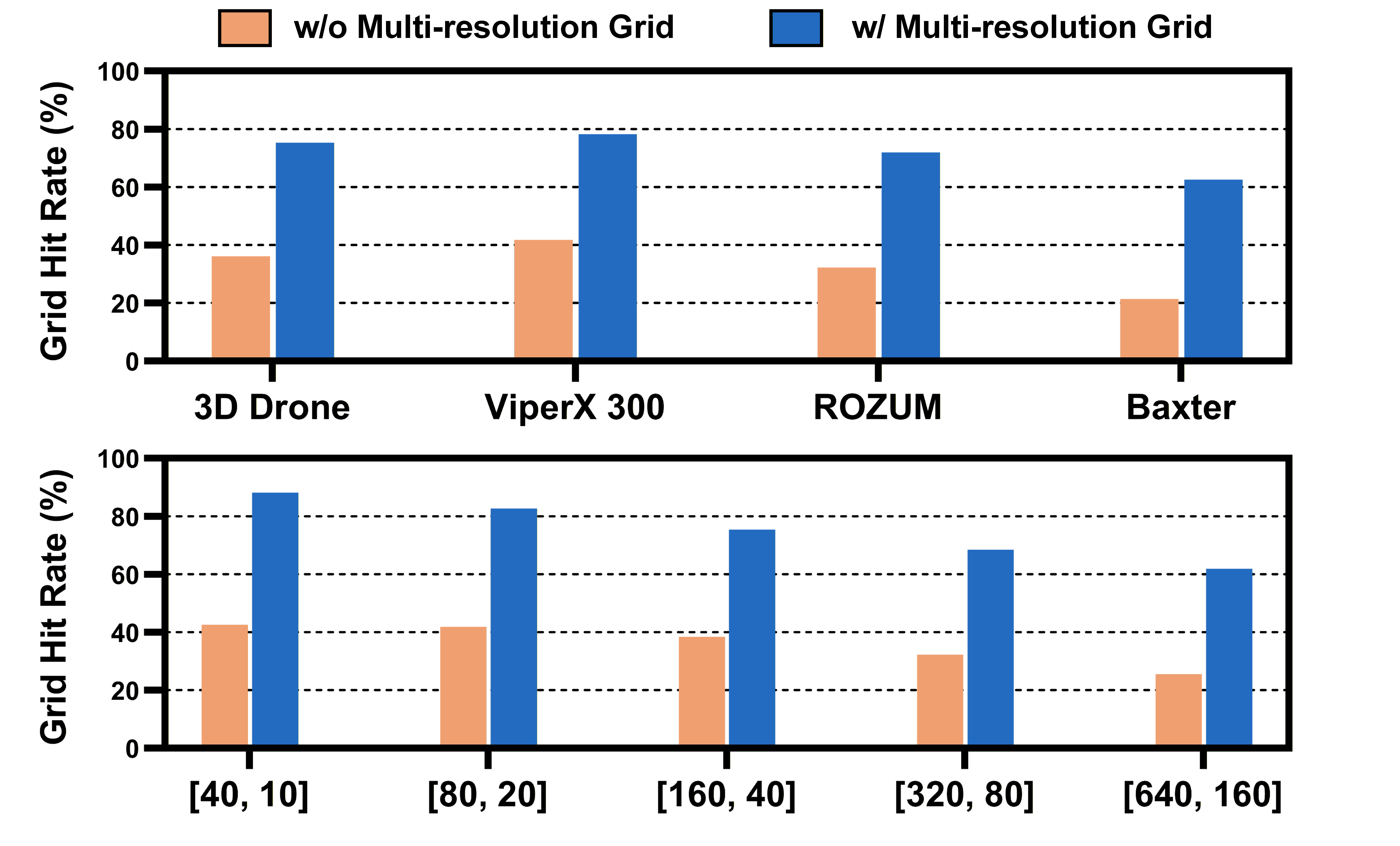}}
\caption{Analysis of multi-resolution grid lookup. (a) Grid hit rate across robot platforms. (b) Grid hit rate across environment scales, denoted as [workspace size, obstacle count].}
\label{img:multi_resolution_hit_rate}
\end{figure}

Fig.~\ref{img:multi_resolution_hit_rate}(a) compares the grid hit
rate across four robot platforms, where multi-resolution lookup
increases the average hit rate from 32.9\% to 72.0\%. Fig.~\ref{img:multi_resolution_hit_rate}(b) shows consistent
improvements as workspace size and obstacle count increase. The average visited-cell count decreases from 124.7 to 4.6, with
the 90th/99th percentiles reduced from 278/574 to 18/27,
confirming that multi-resolution lookup mitigates sparsity-induced
misses and suppresses tail search expansion.

\textbf{Dynamic Memory Reallocation.} To reduce the additional memory cost introduced by multi-resolution storage, we propose a \textit{grid folding} memory management scheme. Initially, a considerable portion of memory is allocated to coarse grids. As planning proceeds, the coarse grids become saturated and the node density exceeds the threshold. Then, their memory is dynamically reclaimed and reallocated to finer grids (Fig.~\ref{img:multi_resolution}(b)). This adaptive strategy maximizes memory reuse across resolutions, compresses overall storage footprint, and sustains high-performance operation.

\subsection{Hash-grid Memory System}
\label{sec:hash_memory}

As identified in Section~\ref{subsec:naive_grid_limits}, the grid representation incurs substantial storage overhead due to sparse occupancy. Many grid cells remain empty but still occupy memory space, while populated cells experience variable occupancy, requiring large fixed-capacity allocations and causing internal fragmentation.

\subsubsection{Hash-based Compression} 
\begin{figure}[!t]
\centerline{\includegraphics[width=0.5\textwidth]{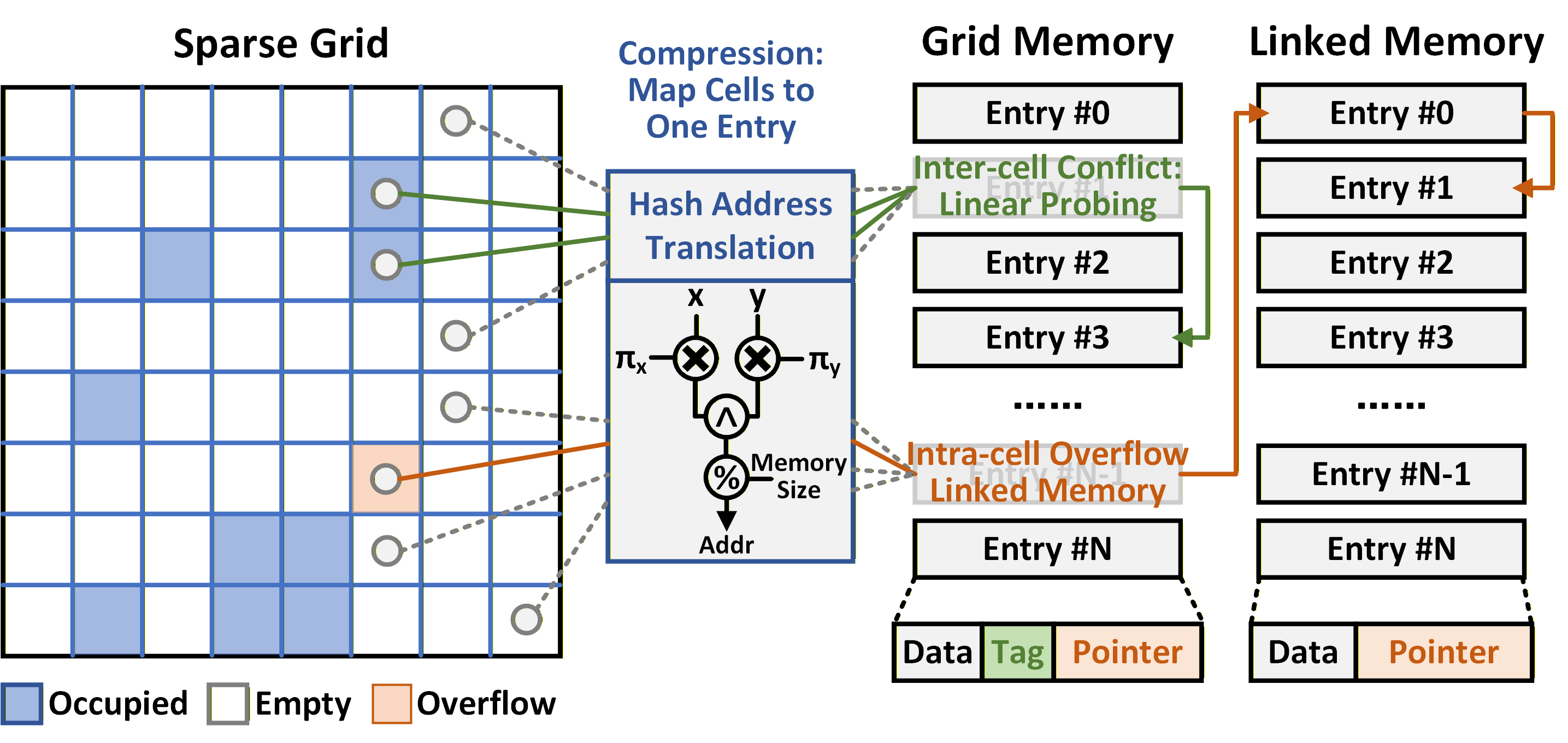}}
\caption{Hash-based compression of a sparse grid into memory. The mapping of grid cells to a compressed Grid Memory is based on hash address translation. The linear probing for inter-cell conflicts and the two-level linked memory for intra-cell overflow enhance its storage efficiency and scalability.}
\label{img:hardware1}
\end{figure}

Fig.~\ref{img:hardware1} gives an overview of the proposed hash-grid memory system. Each grid coordinate $(x,y,z)$ is mapped into a compressed memory space via a simple yet effective hash function: $h(x,y,z) = ((x \cdot p_1 + y \cdot p_2 + z \cdot p_3) \bmod M)$, where $p_1, p_2, p_3$ are large prime numbers and $M$ is the size of the compressed memory space. In the $D$-dimensional case, the formula can be extended as $h(x_1, x_2, ..., x_d) = (\sum x_i \cdot p_i) \bmod M$. This hash function ensures that each address is accessed with equal probability, even under skewed input distributions commonly observed in path planning. The resulting index serves as a direct pointer into a compact hash table, compressing the logical grid space into physical memory.


The sparsity of high-dimensional grids ($>$99.99\% empty cells) enables extreme compression by mapping thousands of logical grid cells to single physical addresses through intentional hash collisions. This approach effectively compresses empty cells, reducing memory footprint by approximately 99.9\% while maintaining direct access to valid cells. Inevitable inter-cell conflicts between occupied logical cells are resolved by fixed-step linear probing, which searches for a hash entry with a matching cell tag or an empty entry for insertion. Empirical analysis shows that the 50th, 90th, and 99th percentile probing depths are 2, 4, and 13, respectively, indicating low common-case access overhead across the evaluated workloads.


\subsubsection{Two-Level Linked Memory} To handle the high variability in cell occupancy (e.g., nodes clustering near obstacles or goal regions), each Grid Memory entry includes a fixed-size local buffer (e.g., 16 node or obstacle IDs) and an overflow pointer to Linked Memory. When the local buffer overflows, a pointer chain links to dynamically allocated chunks in Linked Memory, forming a two-level hybrid structure. Grid Memory stores the entry corresponding to each occupied logical cell, including the cell tag, occupancy metadata (e.g., node/obstacle count and buffer occupancy status), the fixed-size local buffer, and the overflow pointer to Linked Memory. Linked Memory stores only the overflow entries, and pointer chasing is used to traverse these overflow entries. This design achieves three goals: (1) constant-time access in common cases (buffer hit), (2) scalable occupancy support in the worst case (buffer overflow), (3) memory compaction by avoiding pre-allocation for worst-case load. This design achieves a memory footprint less than 0.1\% of a dense grid layout while providing near-constant common-case read and write latency.

\subsection{Pipelined Dataflow and Parallel Scheduling}
\label{sec:dataflow_scheduling}

\begin{figure*}[!t]
\centerline{\includegraphics[width=\textwidth]{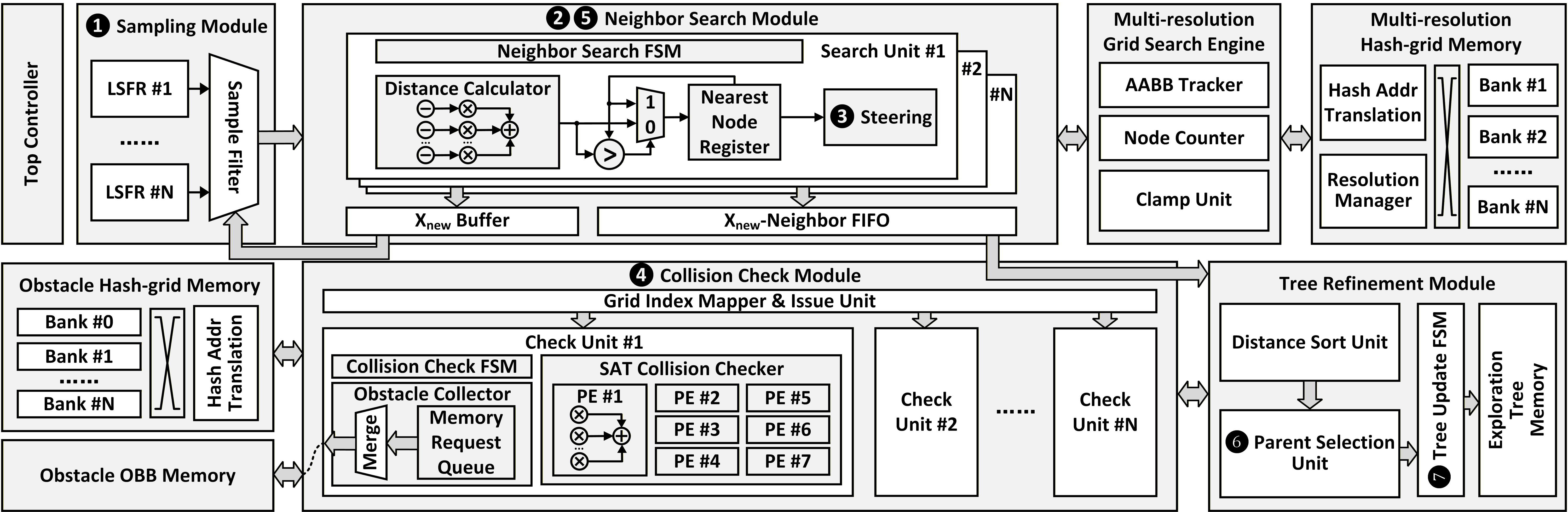}}
\caption{ScaleMPA RRT* accelerator architecture.}
\label{img:hardware_overall}
\end{figure*}

As illustrated in Fig.~\ref{img:hardware_overall}, ScaleMPA comprises five primary functional modules: Sampling Module, Neighbor Search Module, Collision Check Module, Tree Refinement Module and Multi-resolution Grid Search Engine, interacting with four memory subsystems: Exploration Tree Memory, Obstacle Hash-grid Memory, Obstacle OBB Memory and Multi-resolution Hash-grid Memory. 

\subsubsection{Pipeline and Dataflow} The Sampling Module begins with step~\ding{182} (Sampling) by generating multiple random configurations in parallel using a group of Linear Feedback Shift Registers, producing a batch of samples $\{x_{\text{rand}:0}, x_{\text{rand}:1}, \dots, x_{\text{rand}:N-1}\}$. Samples that fall into the same grid cell or are too close to recently generated samples are filtered by the sample filter. The remaining samples are forwarded to the Neighbor Search Module, where multiple samples are processed in parallel through step~\ding{183} (Nearest-Neighbor Search), step~\ding{184} (Steering) and step~\ding{186} (Neighborhood Retrieval). Within the module, grid lookup, candidate collection, distance computation, and steering are organized as buffered pipeline stages to improve hardware utilization under irregular memory access and variable candidate counts.

The Multi-resolution Grid Search Engine retrieves candidate neighbors for each $x_{\text{rand}:i}$ using an adaptively selected grid resolution, after which the nearest node is determined based on Euclidean distance. The steering logic then computes the candidate nodes: $\{x_{\text{new}:0}, x_{\text{new}:1}, \dots, x_{\text{new}:N-1}\}$, by stepping from each \(x_{\text{nearest}:i}\) towards \(x_{\text{rand}:i}\) with a fixed stride. To avoid read and write conflicts, candidates $x_{\text{new}:i}$ falling into the same grid cell are deduplicated.

The Collision Check Module performs step~\ding{185} (Collision Check), validating whether each $x_{\text{new}:i}$ is on a valid, collision-free path. Using grid indexing, obstacle pointers are fetched from the Obstacle Hash-grid Memory, and Obstacle Collector merges identical memory accesses before forwarding them to Obstacle OBB Memory. Corresponding OBB data is then loaded and passed to the SAT Collision Checker for collision verification. The collision-checking datapath further pipelines obstacle-cell generation, obstacle-address collection, OBB loading, and SAT checking. Since different candidate motions may overlap different numbers of obstacles, lightweight buffers are inserted between these stages to absorb latency variations and reduce stalls.

Upon successful collision checks, Neighborhood Retrieval retrieves local neighborhoods
$ \{\mathcal{X}_{\text{near}:0}, \mathcal{X}_{\text{near}:1}, \dots, \mathcal{X}_{\text{near}:N-1}\}$ for each $x_{\text{new}:i}$. Similar to nearest-neighbor search, neighborhood retrieval reuses the grid-lookup and distance-evaluation datapath, and the retrieved neighbor lists are buffered before being consumed by the Tree Refinement Module. The Tree Refinement Module performs step~\ding{187} (Parent Selection) and step~\ding{188} (Rewiring). It evaluates the cost-to-come from all neighbors in $\mathcal{X}_{\text{near}:i}$, selects a valid parent path via collision checking, and commits the result to the Exploration Tree Memory. Rewiring attempts are speculatively executed, with only the first valid connection retained to ensure consistency, as discussed in Section~\ref{sec:parallel_rrt}. The final tree updates are performed through the tree-update logic, which coordinates accepted insertions and rewiring updates before modifying the Exploration Tree Memory.

\subsubsection{Load Balancing and Parallelism Exploration}
To exploit the parallelism enabled by grid representations, ScaleMPA combines sample-level parallelism with fine-grained task-level parallelism. At the sample level, multiple accepted samples in the same batch are processed concurrently after the sample filter removes spatially conflicting samples. This allows independent candidate nodes to proceed through neighbor search, steering, and collision checking in parallel without introducing uncontrolled conflicts in the exploration tree.

At the task level, the workloads of collision checking and neighbor search are highly imbalanced due to spatial variations in obstacle and node densities across grid cells. To prevent straggler-induced bottlenecks, ScaleMPA decomposes the computation for each sample into fine-grained tasks. For collision checking, each candidate motion associated with $x_{\text{new}:i}$ is discretized into multiple spatial configurations, $\{x_{\text{col}:0}, x_{\text{col}:1}, \dots, x_{\text{col}:Q-1}\}$, which are dynamically scheduled across available collision-check units. Similarly, neighbor search and neighborhood retrieval tasks, including grid lookup and pairwise distance computations between $x_{\text{new}:i}$ and nodes in $\mathcal{X}_{\text{near}:i}$, are distributed among parallel execution units. This dynamic scheduling eliminates rigid sample-to-unit mappings, improving throughput and hardware utilization across diverse planning scenarios.

\section{Evaluation}\label{sec:eval}

\subsection{Experimental Methodology}

\textbf{Robotic Systems.} To comprehensively evaluate ScaleMPA, we consider four representative robotic systems with varying degrees of freedom (DOF) and complexity:

\begin{itemize}
  \item \textbf{3D Drone:} A 3-DOF aerial robot with three translational DOF (x, y, z) modeled as a single 3D OBB.
  \item \textbf{ViperX 300 Robot Arm~\cite{viperx}:} A 5-DOF robotic arm with five joint angles, enclosed by three 3D OBBs.
  \item \textbf{ROZUM Robot Arm~\cite{rozum}:} A 6-DOF robotic arm with six joint angles, enclosed by four 3D OBBs.
  \item \textbf{Baxter Robot Arm~\cite{cremer2016performance}:} A 7-DOF robotic arm with seven joint angles, enclosed by seven 3D OBBs.

\end{itemize}


\textbf{Environmental Settings.} We evaluate ScaleMPA under diverse workspace sizes and obstacle densities to assess scalability. Our environment settings follow the commonly used benchmarks~\cite{qureshi2019motion}. Specifically, for 3D navigation tasks, the workspace dimensions range from $40^3$ to $640^3$, and the number of obstacles varies from $10$ to $160$. For robot arm manipulation tasks, the number of obstacles ranges from $5$ to $25$. The default control stride is 2 for drones and 0.1 for robot arms. These settings include environments comparable to those used in prior work~\cite{hpca2024moped} and introduce larger-scale scenarios previously unexplored. For each configuration, we generate $50$ planning tasks by randomly placing obstacles with shapes and sizes consistent with the MPNet environment settings~\cite{qureshi2019motion}. The start and goal configurations for each task are randomly selected to ensure collision-free initialization. All obstacles are represented as OBBs to reflect real-world back-end planners, which typically receive OBB data from perception modules.


\textbf{Baseline.} We evaluate ScaleMPA against:

\begin{itemize}
  \item \textit{CPUs:} C++ implementations of tree-based and grid-based RRT* using RTRBench~\cite{bakhshalipour2022rtrbench} on Intel i7-13650HX CPU. The C++ implementations fully exploit SIMD instructions and are compiled with \texttt{-O3} optimization.
  \item \textit{GPUs:} CUDA-based implementations of tree-based and grid-based RRT* following the massively parallel RRT* design in~\cite{bialkowski2011massively} on an NVIDIA RTX 3090 Ti GPU.
  \item \textit{MPNet Planner~\cite{qureshi2019motion}:} A learning-based motion planner executed on an NVIDIA RTX 3090 Ti GPU, which generates candidate paths using learned heuristics.
  \item \textit{RACOD-based Accelerator~\cite{8056773,bakhshalipour2022racod}:} A hardware RRT* implementation with overlapping tree extension and refinement, and MapReduce-style collision checking.
  \item \textit{MOPED Accelerator~\cite{hpca2024moped}:} A SOTA RRT* accelerator that utilizes MBR-trees for collision checking and neighbor search.
\end{itemize}

For both CPU and GPU baselines, the grid-based RRT* implementation includes software versions of the multi-resolution grid search and hash-based compression proposed in ScaleMPA. This ensures a fair comparison and highlights the performance benefits of the specialized hardware design.

\textbf{Hardware Implementation.} The ScaleMPA hardware prototype is designed in SystemVerilog and synthesized with Synopsys Design Compiler using a commercial 28 nm CMOS library. On-chip SRAM area and energy are estimated using CACTI-7.0. A detailed microarchitectural simulator is used for the behavior modeling. Our simulation framework is also extended to model RACOD and MOPED for fair comparison.

\begin{figure}[!t]
\centerline{\includegraphics[width=0.5\textwidth]{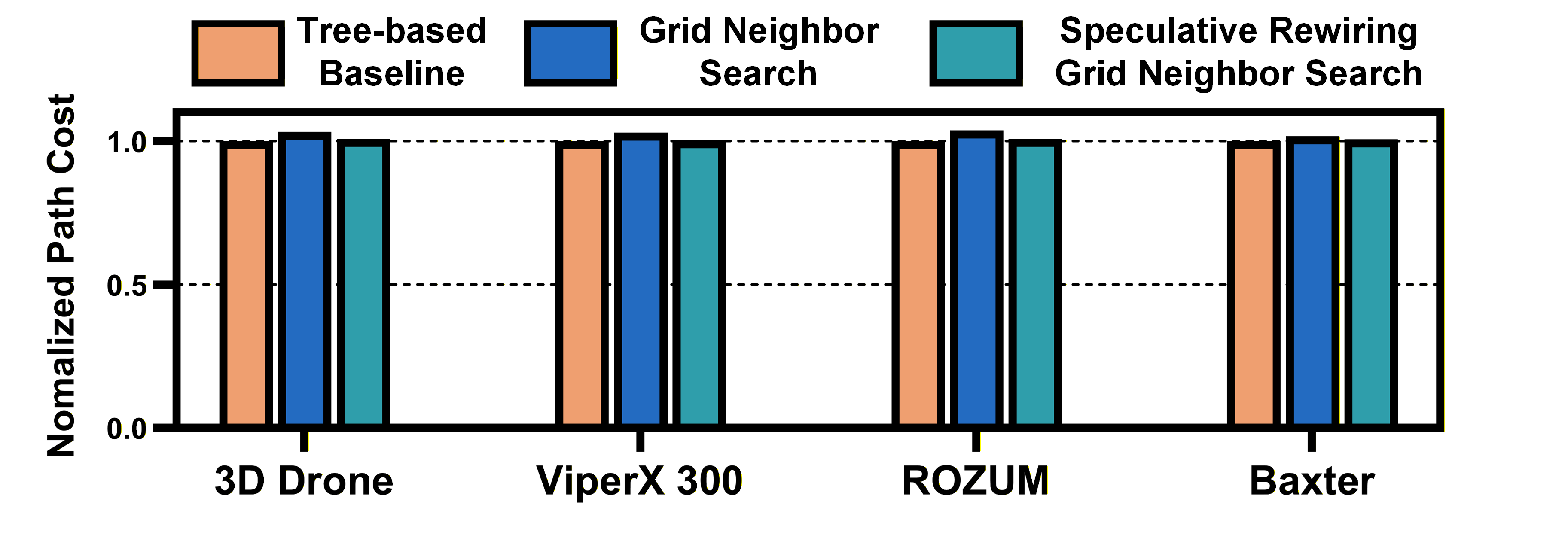}}
\caption{Comparisons on path cost (the lower the better).}
\label{img:eval_algorithm_cost}
\end{figure}

\subsection{Evaluation of Software Implementation}

We evaluate the software performance of the proposed grid-based RRT* algorithm, focusing on its impact on path quality, computational overhead, and empirical scalability.

The ablation study in Fig.~\ref{img:eval_algorithm_cost} evaluates the algorithmic optimality of grid-based neighbor search and cross-iteration parallelism. All the methods use the same sampling sequence, i.e., the same set of sampled configurations to ensure a fair comparison. The results show that grid-based neighbor search has a negligible impact on optimality, while cross-iteration parallelism introduces only a 0.8\% difference in path cost, with identical success rates across all methods, indicating that solution quality and feasibility are preserved.

Based on our empirical evaluation, the grid-based RRT* reduces computational overhead by 97.22\% to 99.44\% over tree-based RRT*. This improvement comes from replacing hierarchical tree traversal with direct grid-based access, which reduces average query cost and yields near-linear end-to-end scaling across the evaluated workloads. This advantage becomes increasingly pronounced as the environment size and obstacle density increase, demonstrating the superior scalability of the grid-based approach. As shown in Fig.~\ref{img:eval_hw_overall}(a), the grid-based approach achieves 1.3$\times$ to 11.3$\times$ speedup over the tree-based implementation on the same CPU, highlighting that the performance gain stems from algorithmic improvements.

\begin{figure}[!t]
\centerline{\includegraphics[width=0.5\textwidth]{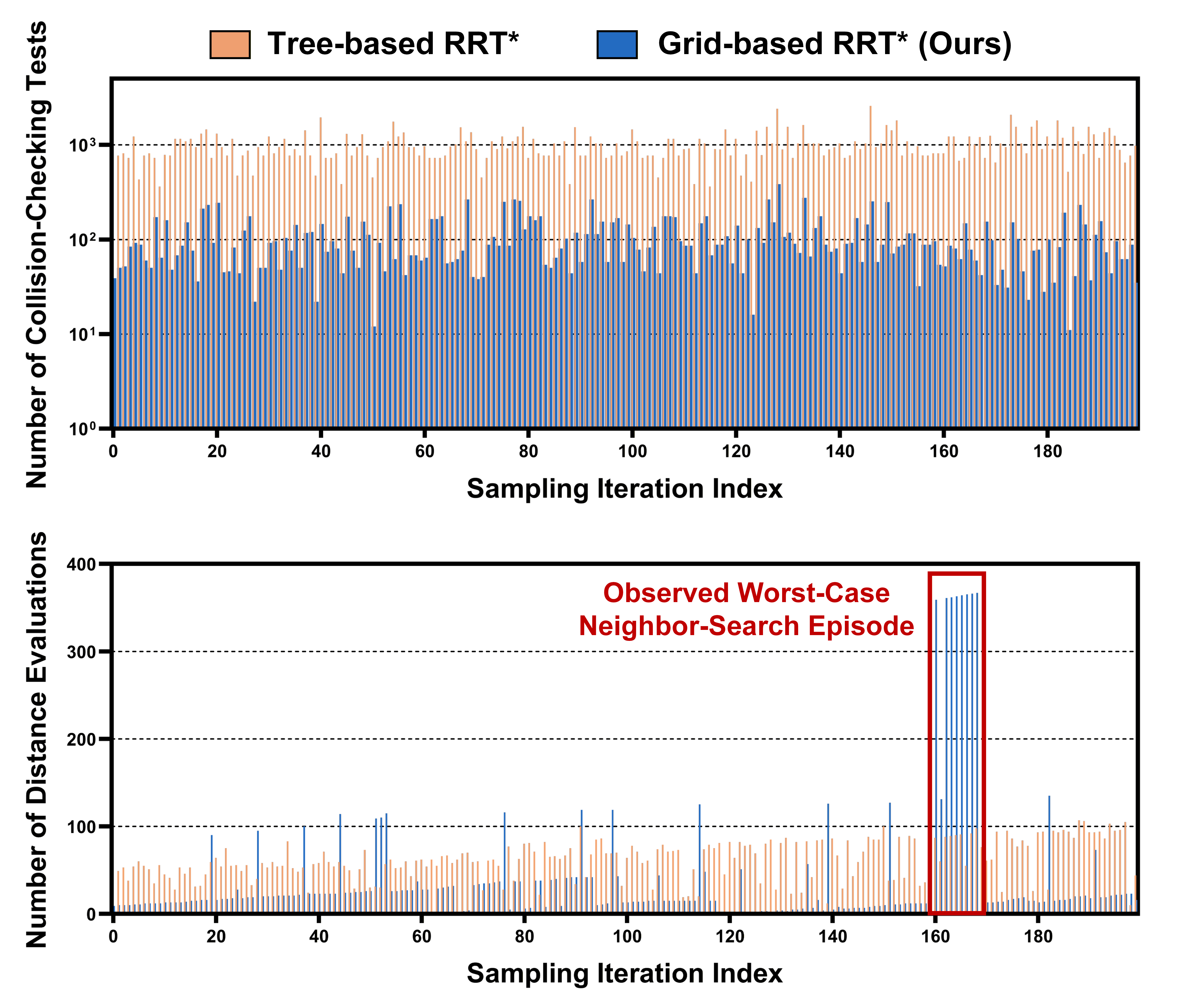}}
\caption{Per-iteration workload comparison of tree-based and grid-based RRT*. (a) Number of collision-checking tests. (b) Number of nearest-neighbor distance evaluations.}
\label{img:worst_case_comparison}
\end{figure}

To examine common-case and worst-case behavior,
Fig.~\ref{img:worst_case_comparison} shows a representative
200-iteration segment from the 3D Drone workload in the largest
evaluated environment, using the same sampling sequence for both methods. As shown in Fig.~\ref{img:worst_case_comparison}(a), grid-based RRT* performs substantially fewer collision-checking tests. In
Fig.~\ref{img:worst_case_comparison}(b), it requires fewer
distance evaluations in most iterations, although the highlighted region reveals a worst-case episode in which repeated misses in empty cells force the search to expand over increasingly larger regions, resulting in a large number of distance evaluations. Consistent with Section~\ref{sec:multi_resolution} and Fig.~\ref{img:multi_resolution_hit_rate}, the 90th/99th-percentile
results show that such cases occur only in the distribution tail.
Thus, despite occasional worst-case penalties, grid-based neighbor
search achieves a substantially lower average computational workload.



\textbf{Key Insights.} (1) Path Quality: Grid-based RRT* preserves solution quality, showing negligible difference in path cost compared to the tree-based baseline. (2) Algorithmic Efficiency: The design offers up to 100$\times$ fewer FLOPs in large maps; despite localized worst-case spikes from miss-driven expansion, its common-case workload remains substantially lower than the tree-based baseline. (3) Scalability: Unlike tree-based methods whose cost increases superlinearly, the grid-based formulation exhibits graceful scaling under increasing workspace size and obstacle density.


\subsection{Evaluation of Hardware Implementation}
\label{sec:eval_hw}

\begin{figure}[!t]
\centerline{\includegraphics[width=0.5\textwidth]{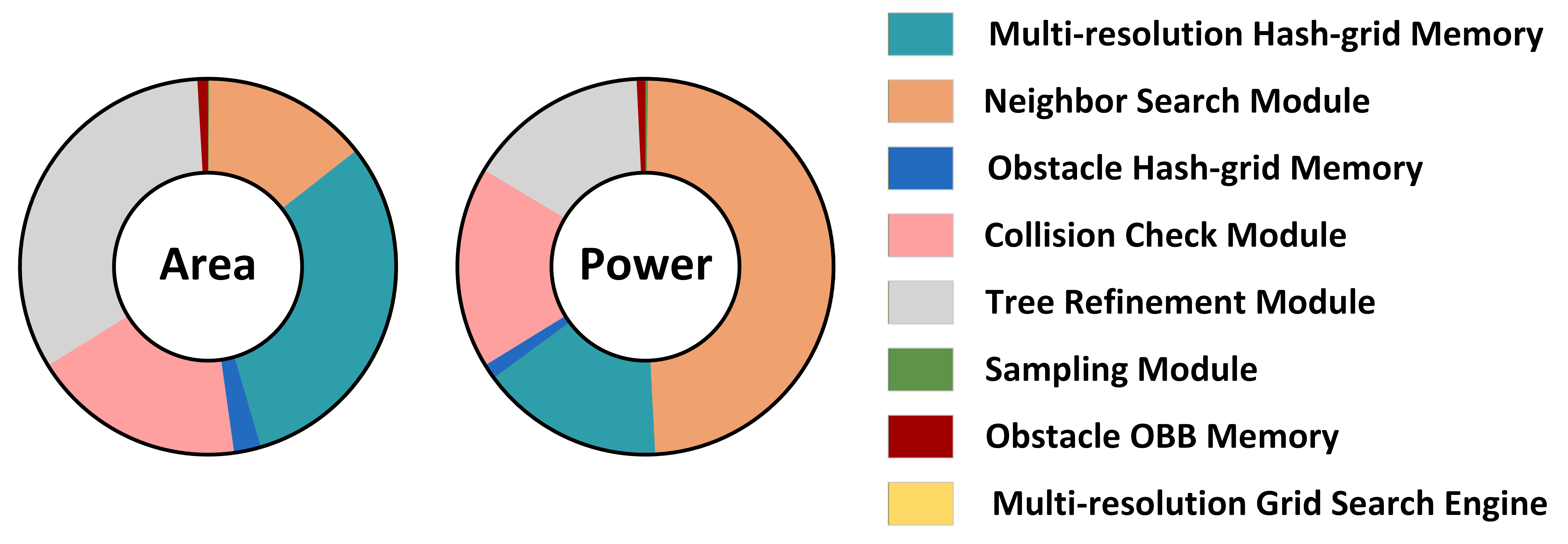}}
\caption{Breakdown of ScaleMPA area and power.}
\label{img:breakdown}
\end{figure}

The ScaleMPA hardware prototype integrates 486 16-bit multipliers, 562 16-bit adders, and 202~KB of on-chip SRAM, using the same numerical precision as MOPED~\cite{hpca2024moped}. It uses a 28 nm CMOS process and achieves an average planning latency of 3.3~ms across all settings, operating at 0.72~V with 89.09~mW power consumption under a 500~MHz clock frequency. The total chip area is 0.88~mm$^2$, with compute logic and on-chip buffers accounting for 33.09\% and 66.91\% of the area, respectively (Fig.~\ref{img:breakdown}). The module-level area distribution is: sampling (0.09\%), neighbor search (14.32\%), collision check (18.42\%), tree refinement (32.89\%), multi-resolution hash-grid memory (31.06\%), obstacle hash-grid memory (2.32\%) and obstacle OBB memory (0.91\%). In terms of power, these components contribute 0.22\%, 48.95\%, 17.40\%, 15.63\%, 15.75\%, 1.30\% and 0.75\%.

\begin{figure*}[!t]
\centerline{\includegraphics[width=\textwidth]{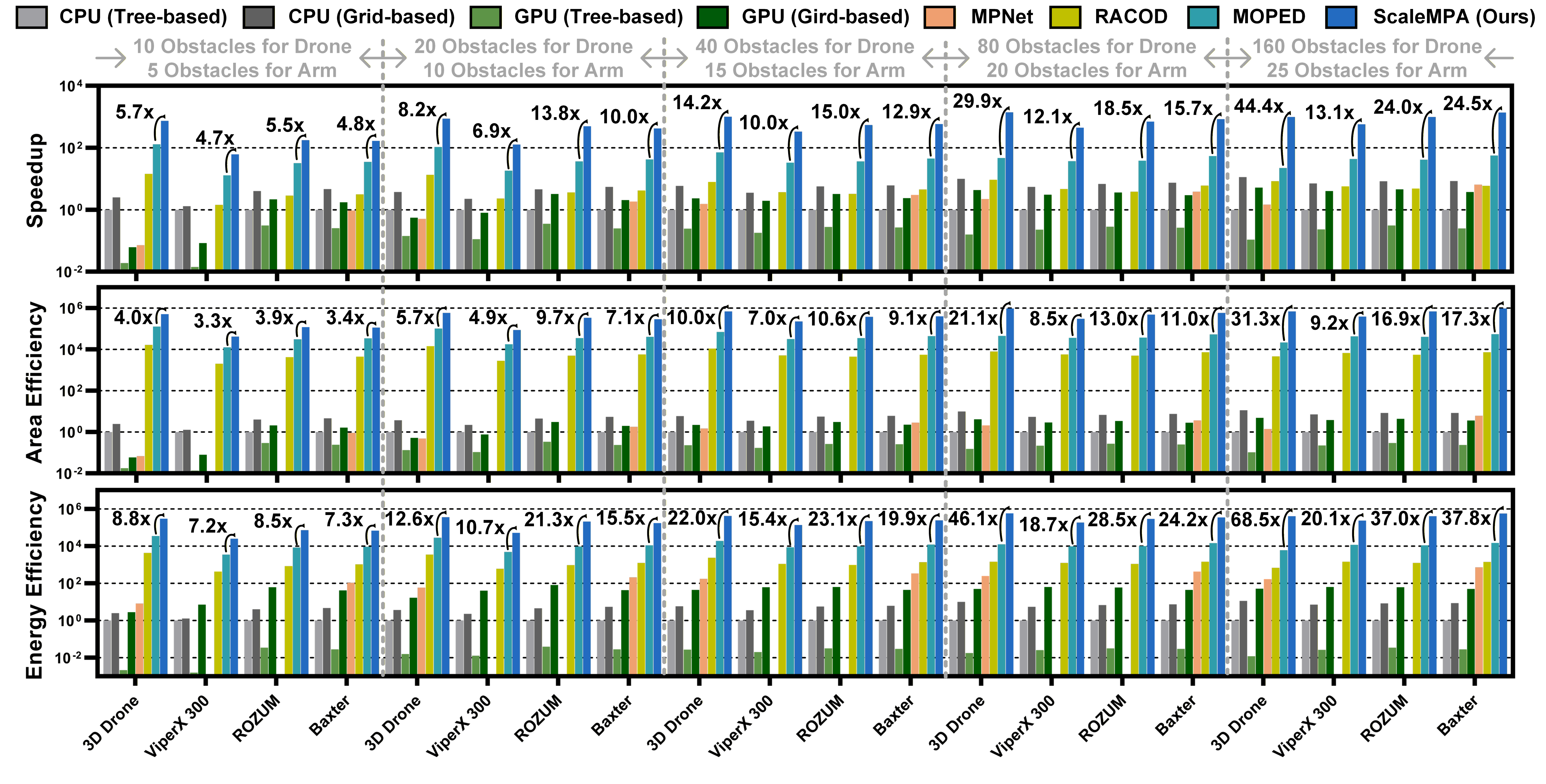}}
\caption{Hardware performance across different robots and environments. Speedup, area efficiency, and energy efficiency are normalized to CPU (Tree-based) baselines. Results cover five environmental settings, comparing CPU (Tree-based and Grid-based), GPU (Tree-based and Grid-based), MPNet, RACOD, MOPED, and ScaleMPA.}
\label{img:eval_hw_overall}
\end{figure*}

Fig.~\ref{img:eval_hw_overall}(a) shows normalized speedups relative to CPU, GPU, MPNet and SOTA accelerators. Since the checkpoints and datasets for ViperX 300 and ROZUM are
unavailable, we report comparisons with MPNet on the 3D Drone and Baxter benchmarks. ScaleMPA outperforms CPU (Tree-based) implementations by 61.1$\times$ to 1372.9$\times$, MPNet by 175.4$\times$ to 10225.2$\times$, and RACOD by 42.7$\times$ to 232.3$\times$. Compared to MOPED, ScaleMPA achieves speedups of 4.7$\times$ to 44.4$\times$. Fig.~\ref{img:eval_hw_overall}(b) and (c) show that ScaleMPA achieves 3.3$\times$–31.3$\times$ higher area efficiency and 7.2$\times$–68.5$\times$ higher energy efficiency than MOPED. Moreover, it delivers average latencies of 3.6\,ms, 1.0\,ms, 4.0\,ms, and 4.6\,ms on the 3D Drone, ViperX 300, ROZUM, and Baxter benchmarks, respectively, with a 99th-percentile iteration latency of 0.018\,ms.


As observed in Section~\ref{subsec:why_not_gpu}, although grid-based RRT* improves parallelism and achieves better GPU utilization than tree-based methods, it still suffers from irregular control flow, thread-level workload imbalance, and non-coalesced memory access, which limit the effectiveness of GPU execution. Consequently, GPU implementations often provide limited benefit and can even underperform optimized CPU versions, as shown in Fig.~\ref{img:eval_hw_overall}(a). In contrast, ScaleMPA translates representation-level parallelism into efficient execution via a pipelined dataflow with fine-grained
scheduling and a hash-grid memory system, enabling substantially higher efficiency and speedup over both CPU and GPU baselines.



\textbf{Key Insights.} (1) Scalability: ScaleMPA sustains low-latency operation across diverse map sizes and obstacle densities, while existing accelerators such as RACOD and MOPED degrade significantly, highlighting the scalability of grid-based RRT*. (2) Hardware efficiency: The architecture achieves high area and energy efficiency by tightly coupling compute-optimized datapaths with memory-aware structures, particularly the multi-resolution and hash-grid memory designs. (3) Generality: Consistent speedups across robotic platforms, from aerial drones to 7-DOF manipulators, demonstrate the architectural generality of ScaleMPA and its applicability to a broad range of motion-planning workloads.



\subsection{Ablation study of Grid-based Acceleration}



\begin{figure}[!t]
\centerline{\includegraphics[width=0.5\textwidth]{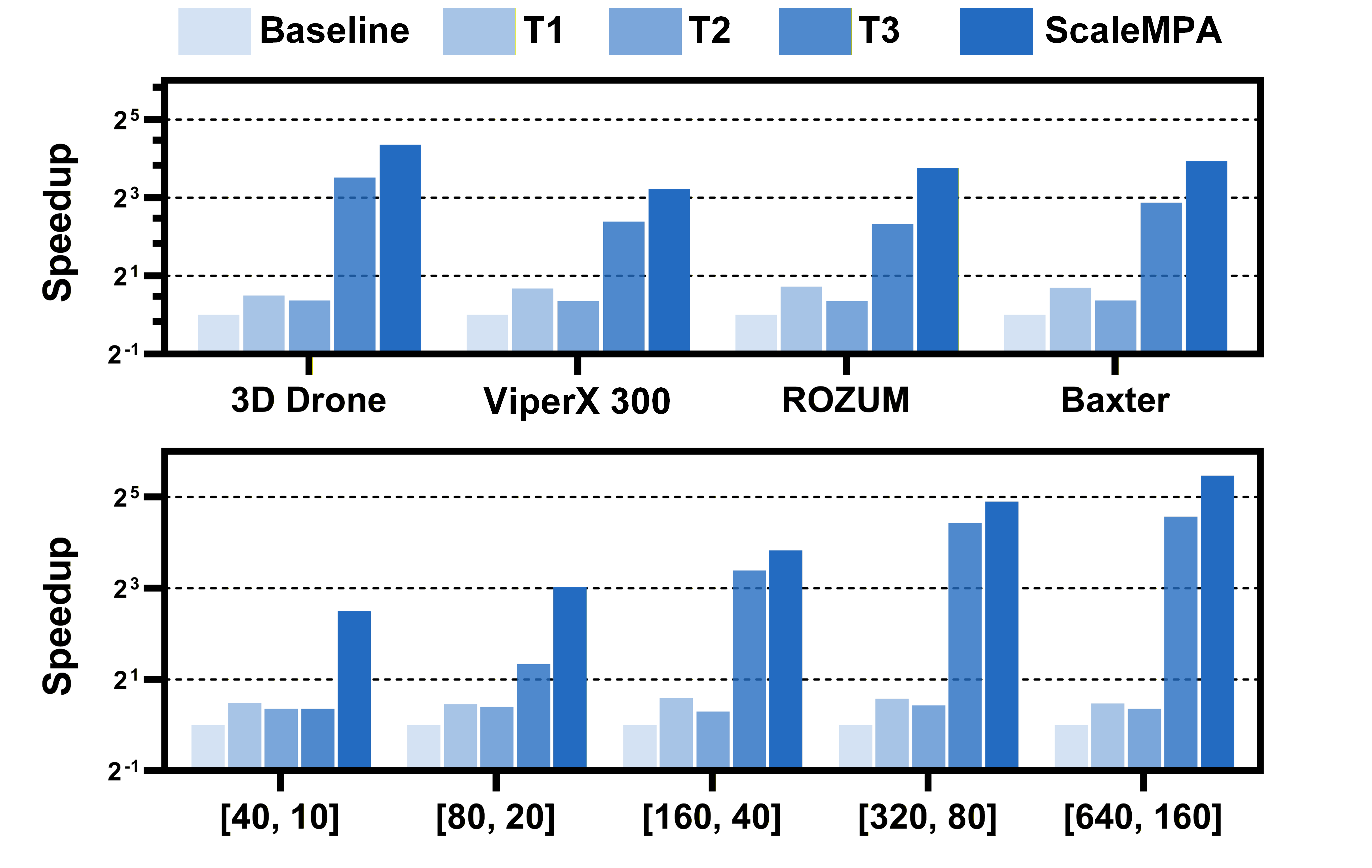}}
\caption{Ablation analysis of grid-based acceleration. (a) Speedup across robot platforms. (b) Speedup across environment scales, denoted as [workspace size, obstacle count]. Baseline: all features disabled; T1: parallel RRT*; T2: grid-based neighbor search; T3: grid-based collision checking; ScaleMPA: full integration of T1–T3.}
\label{img:eval_ablation_study}
\end{figure}

We conduct an ablation study to quantify the benefits of integrating grid-based techniques into different stages of the RRT* hardware architecture, as shown in Fig.~\ref{img:eval_ablation_study}. Five configurations are evaluated: (1) \textbf{Baseline:} all grid-based features disabled; (2) \textbf{T1:} parallel processing of RRT* iterations; (3) \textbf{T2:} grid-based neighbor search; (4) \textbf{T3:} grid-based collision checking; and (5) \textbf{ScaleMPA:} full integration of T1–T3.

As shown in Fig.~\ref{img:eval_ablation_study}(a), all techniques (T1, T2, T3) yield speedups over the baseline. T1 achieves 1.6$\times$ speedup due to parallel iteration processing. T2 adds a 1.3$\times$ improvement, while T3 provides 7.2$\times$ speedup, highlighting that grid-based acceleration is the most effective for collision checking. Finally, the combined ScaleMPA configuration delivers a 14.7$\times$ overall speedup. We also conduct the ablation study under different environment settings to verify the scalability of each technology. Fig.~\ref{img:eval_ablation_study}(b) shows that the benefits of T3 scale with map size and obstacle density. T1 exhibits a constant speedup because the hardware parallelism remains fixed and does not depend on the size of the map. T2 also maintains stable gains due to the grid hit rate.

\subsection{Analysis of Multi-resolution Lookup and Hash-grid Compression}

\begin{figure}[!t]
\centerline{\includegraphics[width=0.5\textwidth]{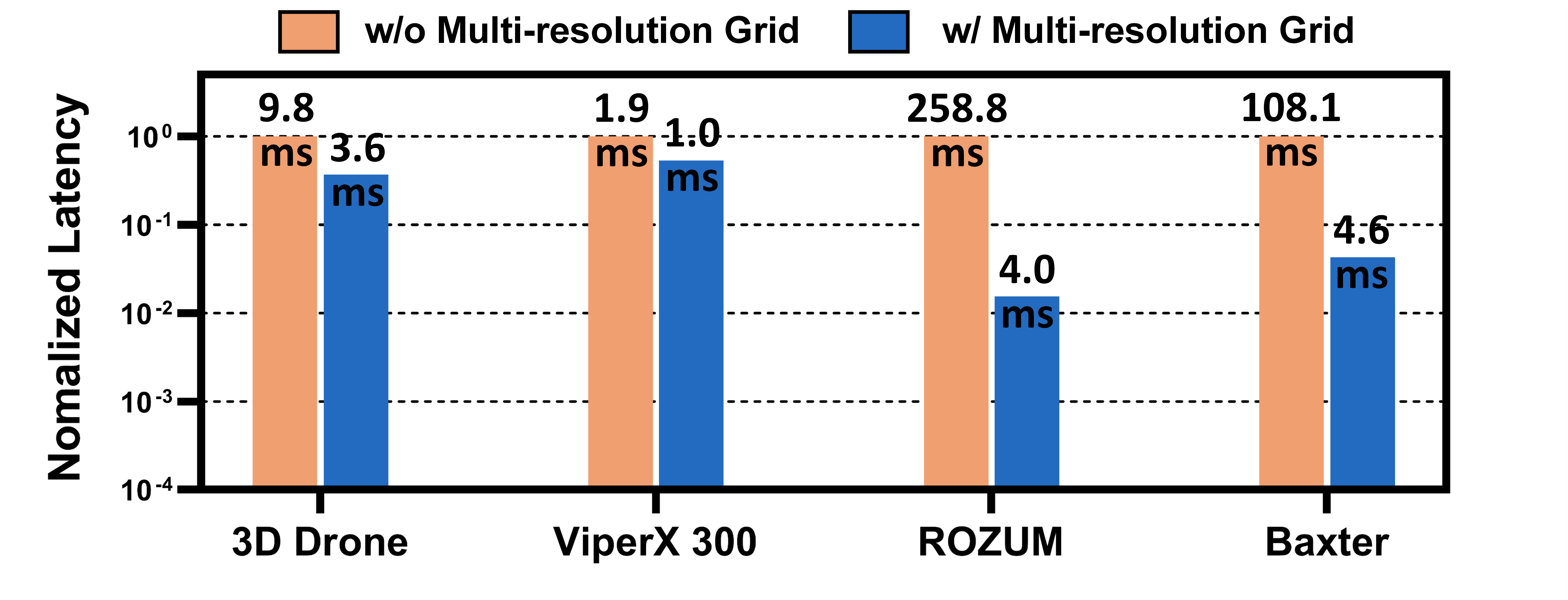}}
\caption{With multi-resolution design, the grid-based RRT* latency is reduced by 46.6\%-98.5\%.}
\label{img:eval_multiresolution}
\end{figure}

As discussed in Section~\ref{sec:multi_resolution}, the multi-resolution grid mitigates the miss penalty inherent in grid-based RRT*. To assess its effectiveness, we compare latency with and without this optimization. Fig.~\ref{img:eval_multiresolution} shows that multi-resolution lookup reduces average latency by 76\%, keeping planning time within the millisecond range. The improvement is especially pronounced for high-DOF robotic arms, where grid sparsity is severe. This confirms that multi-resolution lookup effectively mitigates miss penalties in grid-based RRT*.

\begin{figure}[!t]
\centerline{\includegraphics[width=0.5\textwidth]{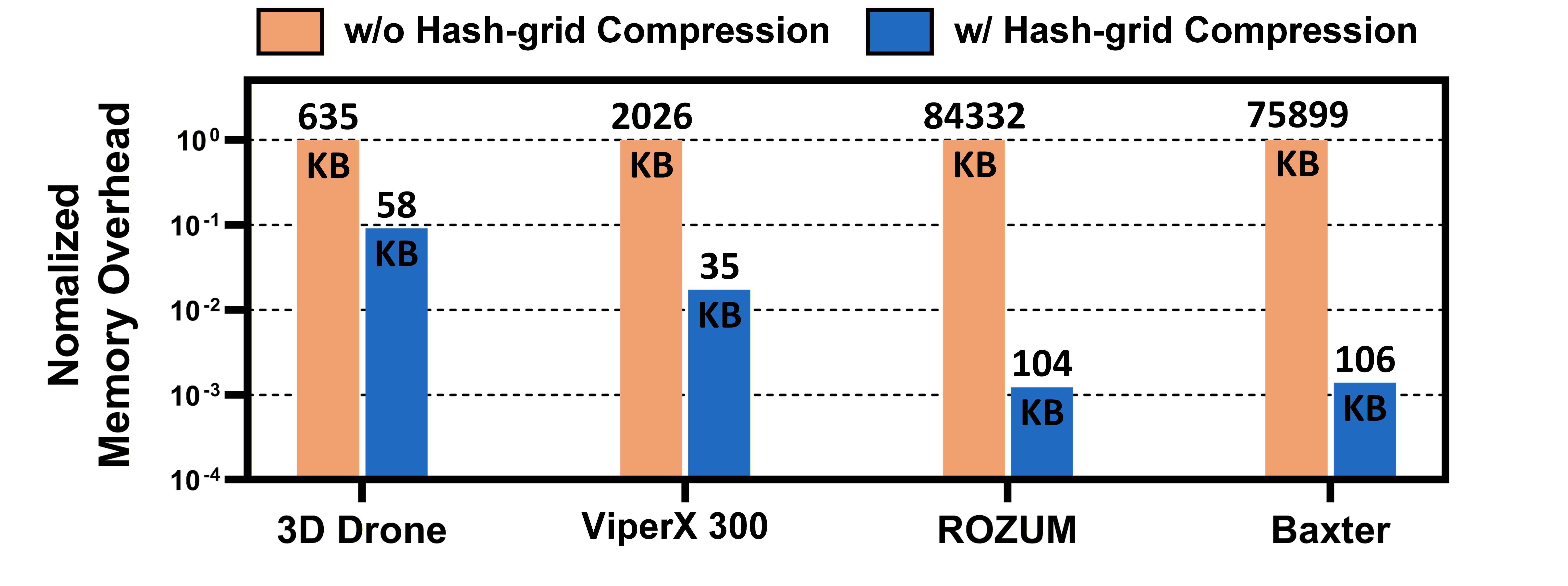}}
\caption{Impact of hash-grid compression on memory overhead. It reduces on-chip memory footprint from 84,332 KB to 106 KB.}
\label{img:eval_hashcompress}
\end{figure}

To evaluate hash-grid compression, we compare the required buffer capacity with and without compression. As illustrated in Fig.~\ref{img:eval_hashcompress}, a naive grid design requires up to 84,332~KB of on-chip memory, which is prohibitively expensive for edge accelerators. By applying hash-grid compression, memory usage drops to less than 106~KB (0.12\%--9.13\% of the original). In addition, we evaluate the load factor and probe depth of the hash structure. The load factor of grid memory ranges from 60\% to 76\% across all settings. The 50th, 90th, and 99th percentile probe depths of grid memory are 2, 4, and 13, respectively, confirming the efficiency of the hash structure.



\subsection{Design Space Exploration}

A design space exploration (DSE) is performed over key parameters, including hardware parallelism, hash-grid memory size, and the number of multi-resolution levels, evaluating trade-offs among latency, area, and energy.

\begin{figure*}[!t]
\centerline{\includegraphics[width=\textwidth]{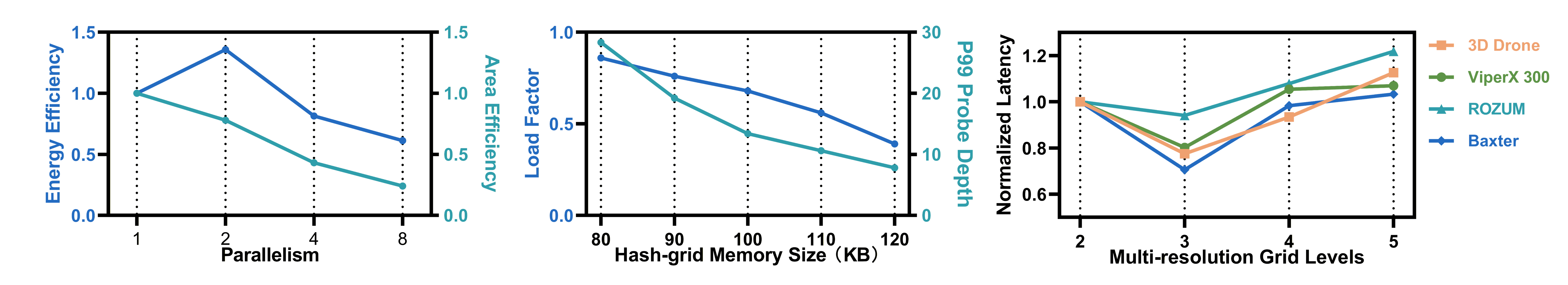}}
\caption{Design space exploration of hardware parallelism, hash-grid memory size and multi-resolution grid levels.}
\label{img:eval_dse}
\end{figure*}

\textbf{Hardware Parallelism:} Increasing parallelism can reduce latency at the cost of higher area and power consumption. As the degree of parallelism increases, the end-to-end latency of RRT* decreases, but area and power overheads grow substantially. To identify the most efficient configuration, we conduct a design space exploration, as shown in Fig.~\ref{img:eval_dse}(a). Among all configurations, a parallelism factor of 2 achieves the highest power efficiency, providing the best trade-off between energy and performance.


\textbf{Hash-grid Memory Size:} The size of the hash-grid memory affects both probing behavior and memory utilization. As shown in Fig.~\ref{img:eval_dse}(b), increasing the memory size reduces the average probing depth, which helps improve lookup latency. However, a larger memory also leads to a lower load factor, resulting in reduced memory utilization efficiency. In addition, increasing the memory capacity incurs higher area and power overhead. We select a hash-grid memory size (106 KB) that balances probing efficiency and hardware utilization, achieving low latency without excessive resource overhead.


\textbf{Multi-resolution Grid Levels:} The number of grid levels directly impacts query latency. As shown in Fig.~\ref{img:eval_dse}(c), increasing the number of grid levels initially reduces latency due to improved spatial partitioning and more efficient neighbor search. However, beyond a certain point, additional grid levels provide diminishing returns and may even introduce overhead. Therefore, we adopt a three-level multi-resolution grid design, which achieves the best trade-off between latency and hardware complexity.


\section{Related Work} \label{sec:related_work}

\textbf{Motion Planning Accelerators.}
Numerous hardware accelerators have been proposed to achieve real-time motion planning. Dadu-P~\cite{dac2018dadup} accelerates PRM-based planning and achieves millisecond-level performance, but lacks optimality guarantees and incurs cubic memory growth with configuration-space size or resolution, limiting its scalability~\cite{bakhshalipour2022racod}. BLITZCRANK~\cite{blitzcrank2023dac} accelerates factor-graph-based planning through incremental inference, achieving 200--300\,Hz for 3-DOF tasks, but its performance drops significantly at higher DOF ($<$10\,Hz at 7-DOF). Dadu-Corki~\cite{huang2025dadu} employs algorithm--architecture co-design for efficient trajectory generation in embodied robotic manipulation, but does not target scalable global exploration in large and complex environments. RACOD~\cite{bakhshalipour2022racod} parallelizes collision detection and look-ahead path search, but remains difficult to scale to high-dimensional spaces. MOPED~\cite{hpca2024moped}, a state-of-the-art RRT*-based accelerator, reduces collision-checking and neighborhood-search complexity using R-tree and MBR-tree structures. Nevertheless, as discussed earlier, tree-based methods still struggle to maintain real-time performance as the configuration space, resolution, or obstacle density increases.


\textbf{Multi-resolution Grid-Based Methods.}
Multi-resolution grid representations have been widely used to improve scalability in large-scale spatial problems. Early works in scientific computing leverage hierarchical grids and adaptive discretization to efficiently solve partial differential equations, where hashing is employed to manage sparse grid structures and enable parallel multigrid solvers~\cite{griebel1998parallel, chow2006survey}. These approaches demonstrate the effectiveness of combining multi-resolution representations with efficient indexing schemes for large-scale problems. However, they are primarily designed for regular numerical workloads and do not directly address the irregular memory access patterns in motion planning, particularly in collision checking and neighbor search.

\textbf{Hashing for Robotics Algorithm Acceleration.}
Hashing has been explored to improve memory efficiency and accelerate spatial queries in robotics. Hash-based spatial indexing reduces collision-detection overhead and improves data locality in robotic accelerators~\cite{shah2024collision}, while spatial hashing supports efficient collision detection and proximity queries in sparse environments~\cite{teschner2003optimized}. Hierarchical hash grids further combine multi-resolution indexing with hash-based storage to reduce broad-phase collision-checking cost and memory usage~\cite{schornbaum2009hierarchical}. However, these approaches mainly target isolated primitives, such as collision prediction, broad-phase filtering, or spatial lookup. Our work integrates hashing with a multi-resolution grid in an end-to-end RRT* accelerator, jointly supporting collision checking, neighbor search, and planner-state updates.

\section{Conclusion}



This paper presents ScaleMPA, an algorithm--architecture co-design for scalable RRT* planning. To overcome the scalability limitations of tree-based RRT*, we replace the tree hierarchy with a grid-native representation that enables direct grid access and exposes more regular parallel work. However, naive grid formulations introduce two challenges: worst-case query degradation and high memory overhead. To address these issues, ScaleMPA incorporates a multi-resolution grid search engine and a hash-grid memory system, reducing average query overhead and exhibiting near-linear end-to-end scaling across the evaluated workloads. Experimental results show that ScaleMPA enables millisecond-level real-time planning in large-scale environments and achieves 4.7$\times$--44.4$\times$ speedup over state-of-the-art accelerators.

\bibliographystyle{IEEEtran}
\bibliography{refs}

@INPROCEEDINGS{8056773,
  author={Xiao, Size and Bergmann, Neil and Postula, Adam},
  booktitle={2017 27th International Conference on Field Programmable Logic and Applications (FPL)}, 
  title={Parallel RRT* architecture design for motion planning}, 
  year={2017},
  volume={},
  number={},
  pages={1-4},
  doi={10.23919/FPL.2017.8056773}}

@article{hart1968formal,
  title={A formal basis for the heuristic determination of minimum cost paths},
  author={Hart, Peter E and Nilsson, Nils J and Raphael, Bertram},
  journal={IEEE transactions on Systems Science and Cybernetics},
  volume={4},
  number={2},
  pages={100--107},
  year={1968},
  publisher={IEEE}
}

@article{ionescu2021adaptive,
  title={Adaptive simplex architecture for safe, real-time robot path planning},
  author={Ionescu, Tudor B},
  journal={Sensors},
  volume={21},
  number={8},
  pages={2589},
  year={2021},
  publisher={MDPI}
}

@INPROCEEDINGS{replan2027iros,
  author={Chandler, Bryant and Goodrich, Michael A.},
  booktitle={2017 IEEE/RSJ International Conference on Intelligent Robots and Systems (IROS)}, 
  title={Online RRT* and online FMT*: Rapid replanning with dynamic cost}, 
  year={2017},
  volume={},
  number={},
  pages={6313-6318},
  doi={10.1109/IROS.2017.8206535}}

@INPROCEEDINGS{fly2019ICRA,
  author={Ryll, Markus and Ware, John and Carter, John and Roy, Nick},
  booktitle={2019 International Conference on Robotics and Automation (ICRA)}, 
  title={Efficient Trajectory Planning for High Speed Flight in Unknown Environments}, 
  year={2019},
  volume={},
  number={},
  pages={732-738},
  doi={10.1109/ICRA.2019.8793930}}

@INPROCEEDINGS{blitzcrank2023dac,
  author={Hao, Yuhui and Gan, Yiming and Yu, Bo and Liu, Qiang and Liu, Shao-Shan and Zhu, Yuhao},
  booktitle={2023 60th ACM/IEEE Design Automation Conference (DAC)}, 
  title={BLITZCRANK: Factor Graph Accelerator for Motion Planning}, 
  year={2023},
  volume={},
  number={},
  pages={1-6},
  doi={10.1109/DAC56929.2023.10247780}}

@inproceedings{bakhshalipour2022rtrbench,
  title={Rtrbench: A benchmark suite for real-time robotics},
  author={Bakhshalipour, Mohammad and Likhachev, Maxim and Gibbons, Phillip B},
  booktitle={2022 IEEE International Symposium on Performance Analysis of Systems and Software (ISPASS)},
  pages={175--186},
  year={2022},
  organization={IEEE}
}

@inproceedings{bakhshalipour2022racod,
  title={Racod: algorithm/hardware co-design for mobile robot path planning},
  author={Bakhshalipour, Mohammad and Ehsani, Seyed Borna and Qadri, Mohamad and Guri, Dominic and Likhachev, Maxim and Gibbons, Phillip B},
  booktitle={Proceedings of the 49th Annual International Symposium on Computer Architecture},
  pages={597--609},
  year={2022}
}

@inproceedings{hpca2024moped,
  title={Moped: Efficient motion planning engine with flexible dimension support},
  author={Huang, Lingyi and Gong, Yu and Sui, Yang and Zang, Xiao and Yuan, Bo},
  booktitle={2024 IEEE International Symposium on High-Performance Computer Architecture (HPCA)},
  pages={483--497},
  year={2024},
  organization={IEEE}
}

@inproceedings{dac2018dadup,
  title={Dadu-p: A scalable accelerator for robot motion planning in a dynamic environment},
  author={Lian, Shiqi and Han, Yinhe and Chen, Xiaoming and Wang, Ying and Xiao, Hang},
  booktitle={Proceedings of the 55th Annual Design Automation Conference},
  pages={1--6},
  year={2018}
}

@article{ijrr2012gpucc,
  title={GPU-based parallel collision detection for fast motion planning},
  author={Pan, Jia and Manocha, Dinesh},
  journal={The International Journal of Robotics Research},
  volume={31},
  number={2},
  pages={187--200},
  year={2012},
  publisher={SAGE Publications Sage UK: London, England}
}

@inproceedings{isca2023mp,
  title={Energy-efficient realtime motion planning},
  author={Shah, Deval and Yang, Ningfeng and Aamodt, Tor M},
  booktitle={Proceedings of the 50th Annual International Symposium on Computer Architecture},
  pages={1--17},
  year={2023}
}

@inproceedings{dac2020dadu,
  title={Dadu-CD: Fast and efficient processing-in-memory accelerator for collision detection},
  author={Yang, Yuxin and Chen, Xiaoming and Han, Yinhe},
  booktitle={2020 57th ACM/IEEE Design Automation Conference (DAC)},
  pages={1--6},
  year={2020},
  organization={IEEE}
}

@inproceedings{hpca2023parallelnn,
  title={Parallelnn: A parallel octree-based nearest neighbor search accelerator for 3d point clouds},
  author={Chen, Faquan and Ying, Rendong and Xue, Jianwei and Wen, Fei and Liu, Peilin},
  booktitle={2023 IEEE International Symposium on High-Performance Computer Architecture (HPCA)},
  pages={403--414},
  year={2023},
  organization={IEEE}
}

@inproceedings{hpca2020quicknn,
  title={Quicknn: Memory and performance optimization of kd tree based nearest neighbor search for 3d point clouds},
  author={Pinkham, Reid and Zeng, Shuqing and Zhang, Zhengya},
  booktitle={2020 IEEE International symposium on high performance computer architecture (HPCA)},
  pages={180--192},
  year={2020},
  organization={IEEE}
}

@inproceedings{sig2022rtnn,
  title={RTNN: accelerating neighbor search using hardware ray tracing},
  author={Zhu, Yuhao},
  booktitle={Proceedings of the 27th ACM SIGPLAN Symposium on Principles and Practice of Parallel Programming},
  pages={76--89},
  year={2022}
}

@inproceedings{strub2020advanced,
  title={Advanced BIT*(ABIT*): Sampling-based planning with advanced graph-search techniques},
  author={Strub, Marlin P and Gammell, Jonathan D},
  booktitle={2020 IEEE International Conference on Robotics and Automation (ICRA)},
  pages={130--136},
  year={2020},
  organization={IEEE}
}

@inproceedings{ortiz2024idb,
  title={idb-rrt: Sampling-based kinodynamic motion planning with motion primitives and trajectory optimization},
  author={Ortiz-Haro, Joaquim and H{\"o}nig, Wolfgang and Hartmann, Valentin N and Toussaint, Marc and Righetti, Ludovic},
  booktitle={2024 IEEE/RSJ International Conference on Intelligent Robots and Systems (IROS)},
  pages={10702--10709},
  year={2024},
  organization={IEEE}
}

@inproceedings{thomason2024motions,
  title={Motions in microseconds via vectorized sampling-based planning},
  author={Thomason, Wil and Kingston, Zachary and Kavraki, Lydia E},
  booktitle={2024 IEEE International Conference on Robotics and Automation (ICRA)},
  pages={8749--8756},
  year={2024},
  organization={IEEE}
}

@article{safaoui2024safe,
  title={Safe multiagent motion planning under uncertainty for drones using filtered reinforcement learning},
  author={Safaoui, Sleiman and Vinod, Abraham P and Chakrabarty, Ankush and Quirynen, Rien and Yoshikawa, Nobuyuki and Di Cairano, Stefano},
  journal={IEEE Transactions on Robotics},
  volume={40},
  pages={2529--2542},
  year={2024},
  publisher={IEEE}
}

@article{xie2022distributed,
  title={Distributed motion planning for safe autonomous vehicle overtaking via artificial potential field},
  author={Xie, Songtao and Hu, Junyan and Bhowmick, Parijat and Ding, Zhengtao and Arvin, Farshad},
  journal={IEEE Transactions on Intelligent Transportation Systems},
  volume={23},
  number={11},
  pages={21531--21547},
  year={2022},
  publisher={IEEE}
}

@article{orthey2023sampling,
  title={Sampling-based motion planning: A comparative review},
  author={Orthey, Andreas and Chamzas, Constantinos and Kavraki, Lydia E},
  journal={Annual Review of Control, Robotics, and Autonomous Systems},
  volume={7},
  year={2023},
  publisher={Annual Reviews}
}

@article{de2024humanoid,
  title={Humanoid robot motion planning approaches: a survey},
  author={de Lima, Carolina Rutili and Khan, Said G and Tufail, Muhammad and Shah, Syed H and Maximo, Marcos ROA},
  journal={Journal of Intelligent \& Robotic Systems},
  volume={110},
  number={2},
  pages={86},
  year={2024},
  publisher={Springer}
}

@misc{rozum,
  author       = {R. Robotics},
  title        = {Roboticarms pulse},
  howpublished = {\url{https://rozum.com/robotic-arm/#about}},
  note         = {Online; accessed 2023-07-24}
}

@misc{viperx,
  author       = {T. Robotics},
  title        = {Viperx 300 robot arm},
  howpublished = {\url{https://www.trossenrobotics.com/viperx-300-robot-arm.aspx}},
  note         = {Online; accessed 2023-07-24}
}

@article{bjorck2025gr00t,
  title={Gr00t n1: An open foundation model for generalist humanoid robots},
  author={Bjorck, Johan and Casta{\~n}eda, Fernando and Cherniadev, Nikita and Da, Xingye and Ding, Runyu and Fan, Linxi and Fang, Yu and Fox, Dieter and Hu, Fengyuan and Huang, Spencer and others},
  journal={arXiv preprint arXiv:2503.14734},
  year={2025}
}

@inproceedings{ratliff2009chomp,
  title={CHOMP: Gradient optimization techniques for efficient motion planning},
  author={Ratliff, Nathan and Zucker, Matt and Bagnell, J Andrew and Srinivasa, Siddhartha},
  booktitle={2009 IEEE international conference on robotics and automation},
  pages={489--494},
  year={2009},
  organization={IEEE}
}

@article{schulman2014motion,
  title={Motion planning with sequential convex optimization and convex collision checking},
  author={Schulman, John and Duan, Yan and Ho, Jonathan and Lee, Alex and Awwal, Ibrahim and Bradlow, Henry and Pan, Jia and Patil, Sachin and Goldberg, Ken and Abbeel, Pieter},
  journal={The International Journal of Robotics Research},
  volume={33},
  number={9},
  pages={1251--1270},
  year={2014},
  publisher={Sage Publications Sage UK: London, England}
}

@article{karaman2011sampling,
  title={Sampling-based algorithms for optimal motion planning},
  author={Karaman, Sertac and Frazzoli, Emilio},
  journal={The international journal of robotics research},
  volume={30},
  number={7},
  pages={846--894},
  year={2011},
  publisher={Sage Publications Sage UK: London, England}
}

@inproceedings{qureshi2019motion,
  title={Motion planning networks},
  author={Qureshi, Ahmed H and Simeonov, Anthony and Bency, Mayur J and Yip, Michael C},
  booktitle={2019 International Conference on Robotics and Automation (ICRA)},
  pages={2118--2124},
  year={2019},
  organization={IEEE}
}

@article{latif20243p,
  title={3p-llm: Probabilistic path planning using large language model for autonomous robot navigation},
  author={Latif, Ehsan},
  journal={arXiv preprint arXiv:2403.18778},
  year={2024}
}

@inproceedings{meng2024llm,
  title={Llm-a*: Large language model enhanced incremental heuristic search on path planning},
  author={Meng, Silin and Wang, Yiwei and Yang, Cheng-Fu and Peng, Nanyun and Chang, Kai-Wei},
  booktitle={Findings of the Association for Computational Linguistics: EMNLP 2024},
  pages={1087--1102},
  year={2024}
}

@article{kim2024openvla,
  title={Openvla: An open-source vision-language-action model},
  author={Kim, Moo Jin and Pertsch, Karl and Karamcheti, Siddharth and Xiao, Ted and Balakrishna, Ashwin and Nair, Suraj and Rafailov, Rafael and Foster, Ethan and Lam, Grace and Sanketi, Pannag and others},
  journal={arXiv preprint arXiv:2406.09246},
  year={2024}
}

@article{black2024pi_0,
  title={$\pi_0$: A Vision-Language-Action Flow Model for General Robot Control},
  author={Black, Kevin and Brown, Noah and Driess, Danny and Esmail, Adnan and Equi, Michael and Finn, Chelsea and Fusai, Niccolo and Groom, Lachy and Hausman, Karol and Ichter, Brian and others},
  journal={arXiv preprint arXiv:2410.24164},
  year={2024}
}

@inproceedings{cremer2016performance,
  title={On the performance of the Baxter research robot},
  author={Cremer, Sven and Mastromoro, Lawrence and Popa, Dan O},
  booktitle={2016 IEEE international symposium on assembly and manufacturing (ISAM)},
  pages={106--111},
  year={2016},
  organization={IEEE}
}

@inproceedings{bialkowski2011massively,
  title={Massively parallelizing the RRT and the RRT*},
  author={Bialkowski, Joshua and Karaman, Sertac and Frazzoli, Emilio},
  booktitle={2011 IEEE/RSJ International Conference on Intelligent Robots and Systems},
  pages={3513--3518},
  year={2011},
  organization={IEEE}
}

@incollection{griebel1998parallel,
  title={Parallel multigrid in an adaptive PDE solver based on hashing},
  author={Griebel, Michael and Zumbusch, Gerhard},
  booktitle={Advances in Parallel Computing},
  volume={12},
  pages={589--599},
  year={1998},
  publisher={Elsevier}
}

@article{chow2006survey,
  title={A survey of parallelization techniques for multigrid solvers},
  author={Chow, Edmond and Falgout, Robert D and Hu, Jonathan J and Tuminaro, Raymond S and Yang, Ulrike Meier},
  journal={Parallel processing for scientific computing},
  pages={179--201},
  year={2006},
  publisher={SIAM}
}

@inproceedings{shah2024collision,
  title={Collision Prediction for Robotics Accelerators},
  author={Shah, Deval and Aamodt, Tor M},
  booktitle={2024 ACM/IEEE 51st Annual International Symposium on Computer Architecture (ISCA)},
  pages={566--581},
  year={2024},
  organization={IEEE}
}

@inproceedings{teschner2003optimized,
  title={Optimized spatial hashing for collision detection of deformable objects.},
  author={Teschner, Matthias and Heidelberger, Bruno and M{\"u}ller, Matthias and Pomerantes, Danat and Gross, Markus H},
  booktitle={Vmv},
  volume={3},
  pages={47--54},
  year={2003}
}

@article{schornbaum2009hierarchical,
  title={Hierarchical hash grids for coarse collision detection},
  author={Schornbaum, Florian},
  journal={Student Thesis, University of Erlangen-Nuremberg},
  year={2009}
}

@inproceedings{huang2025dadu,
  title={Dadu-corki: Algorithm-architecture co-design for embodied ai-powered robotic manipulation},
  author={Huang, Yiyang and Hao, Yuhui and Yu, Bo and Yan, Feng and Yang, Yuxin and Min, Feng and Han, Yinhe and Ma, Lin and Liu, Shaoshan and Liu, Qiang and others},
  booktitle={Proceedings of the 52nd Annual International Symposium on Computer Architecture},
  pages={327--343},
  year={2025}
}

@inproceedings{wilson2025nearest,
  title={Nearest-neighbourless asymptotically optimal motion planning with Fully Connected Informed Trees (FCIT*)},
  author={Wilson, Tyler S and Thomason, Wil and Kingston, Zachary and Kavraki, Lydia E and Gammell, Jonathan D},
  booktitle={2025 IEEE International Conference on Robotics and Automation (ICRA)},
  pages={14140--14146},
  year={2025},
  organization={IEEE}
}

@article{huang2025prrtc,
  title={prrtc: Gpu-parallel rrt-connect for fast, consistent, and low-cost motion planning},
  author={Huang, Chih H and Jadhav, Pranav and Plancher, Brian and Kingston, Zachary},
  journal={arXiv preprint arXiv:2503.06757},
  year={2025}
}

 

\makeatletter
\def\@IEEEBIOskipN{1\baselineskip}
\makeatother

\vfill

\end{document}